\documentclass{article}
\pdfoutput=1
\usepackage{arxiv}

\usepackage[utf8]{inputenc} 
\usepackage[T1]{fontenc}    
\usepackage{hyperref}       
\usepackage{url}            
\usepackage{booktabs}       
\usepackage{amsmath,amssymb,amsfonts} 
\usepackage{nicefrac}       
\usepackage{microtype}      
\usepackage{cleveref}       
\usepackage{graphicx}
\usepackage{natbib}
\usepackage{doi}
\usepackage{subcaption}
\usepackage{multirow}
\usepackage{float}
\usepackage{textcomp}
\usepackage{xcolor}
\usepackage{algorithm}
\usepackage{algorithmic}
\usepackage{needspace}

\title{PhenoNEST: A Neuro-Symbolic Framework for Ontology-Aware Multimodal Plant Phenotyping and Trait Discovery}

\newif\ifuniqueAffiliation
\uniqueAffiliationtrue

\ifuniqueAffiliation
\author{
  Jayant Ghadge \\
  Centre of Studies in Resources Engineering (CSRE) \\
  Indian Institute of Technology Bombay \\
  Mumbai, India \\
  \And
  Soumyashree Kar\thanks{Corresponding author} \\
  Centre of Studies in Resources Engineering (CSRE) \\
  Indian Institute of Technology Bombay \\
  Mumbai, India \\
  \texttt{soumyakar@iitb.ac.in} \\
  \And
  Surya S. Durbha \\
  Centre of Studies in Resources Engineering (CSRE) \\
  Indian Institute of Technology Bombay \\
  Mumbai, India \\
}
\fi

\renewcommand{\shorttitle}{PhenoNEST}

\hypersetup{
pdftitle={PhenoNEST: A Neuro-Symbolic Framework for Ontology-Aware Multimodal Plant Phenotyping and Trait Discovery},
pdfsubject={cs.AI, cs.CV},
pdfauthor={Jayant Ghadge, Soumyashree Kar, Surya S. Durbha},
pdfkeywords={Multi-modal AI, Vision-Language Models, Knowledge Graphs, Precision Agriculture, Visual Grounding},
}

\begin{document}
\maketitle

\begin{abstract}
High-throughput plant phenotyping generates valuable data that often remains trapped in unstructured text and isolated RGB images. To bridge this semantic gap, we propose a framework for constructing a multimodal granular Knowledge Graph (KG) to monitor genotype-phenotype interactions across time and experiments. In this work, we focus on wheat (\textit{Triticum aestivum}) as a representative target crop to validate our methodology across complex canopy environments. Our pipeline first distills noisy field notes to extract entities and relations, dynamically constructing the KG by converting unique instances into hierarchical class entities via RDF-typing. These graph nodes are then aligned with standardized ontologies (PO, RO, WTO) using PlantDeBERTa. To visually ground the constructed graph, a Vision-Language Model paired with a wheat-segmentation ViT generates attention-based softmaps, linking specific KG entities directly to image pixels. We introduce a central observation node (\texttt{Plant\_Obs\_Id}) to connect these multimodal subgraphs temporally. Evaluated on 500 curated WisWheat samples using Pointing Game accuracy, Visual Word Sense Disambiguation (VWSD), and rank-based metrics, our neuro-symbolic approach successfully maps complex field observations to a structured graph. This enables automated field note auditing, temporal stress monitoring, and precise spatial trait localization for wheat breeders.
\end{abstract}


\section{Introduction}
\label{sec:introduction}

High-throughput plant phenotyping (HTP) has witnessed significant acceleration due to advancements in computer vision and Artificial Intelligence (AI). While HTP expedites plant observations and analytics in streamlined workflows—often autonomous and automated—these systems rely heavily on multi-source or multi-modal data inputs to capture the true multidimensionality of competing plant traits and processes. This complexity escalates significantly in field environments compared to controlled laboratory settings. For non-controlled fields prone to drastic ambient variations in the environment, multimodal platforms are essential for simultaneous data collection. These visual streams are then integrated with unstructured field notes from human experts to provide credible, relevant, and ground-truthed solutions to domain stakeholders, such as plant breeders and agricultural scientists. 

However, standard architectures treat these multi-modal streams (such as field images and textual notes) in isolation or rely on surface-level visual patterns. They remain overwhelmingly vision-dominant and are led by traditional vision models like the Vision Transformer (ViT), ResNet, and EfficientNet \cite{resnet2016, efficientnet2019, vit2021} rather than explicit biological definitions. Because these stand-alone systems rely on purely statistical correlations, they fail to provide interpretable or explainable predictions, especially regarding trait associations and trait discovery. Black-box deep learning layers excel at spotting low-level pixel patterns, but they cannot inherently map a visual anomaly to its formal pathological or physiological definition (e.g., distinguishing early chlorosis from natural leaf senescence). Without a structured framework, these models cannot automatically connect related traits over time, trace a visible symptom back to its true underlying biotic or abiotic causes, or provide a transparent decision trail that a breeder can trust for scientific validation.

To resolve these limitations, modern HTP demands a tight conceptual linkage between state-of-the-art Vision-Language Models and established, structured domain knowledge. Integrating formal biological ontologies ensures that raw, multi-source agronomic observations are mapped into a standardized, interoperable, and completely unambiguous semantic structure. By anchoring visual features directly to standardized biological definitions, the AI system transcends surface-level patterns to achieve true biological context awareness. Furthermore, leveraging these structured classification hierarchies enables the system to support flexible, natural language query interfaces. Instead of writing rigid database code, researchers can query the system using intuitive, unconstrained language, allowing the framework to dynamically navigate complex crop and disease taxonomies behind the scenes and instantly retrieve grounded visual evidence.

To establish this rigorous foundation for automated trait discovery and longitudinal crop monitoring, this paper introduces \textit{PhenoNest}, a framework designed to bridge the gap between high-resolution field imagery and unstructured biological field notes. By separating image perception from biological reasoning logic, our framework constructs a Multimodal Knowledge Graph (MKG) a highly structured, searchable database network where every visual trait is explicitly anchored to standardized biological schemas and physical image coordinates.
\section{Related Work}
\label{sec:related_work}
The integration of multi-modal AI systems in precision agriculture requires a convergence of vision-language processing, automated language agents, textual entity linking, and structural knowledge graph modeling. Following are some works that are relevant to this field :
\begin{itemize}
\item \textbf{Vision-Language Modeling in Agronomy}
Recent benchmarks have focused heavily on fine-tuning open-source vision backbones for complex open-ended agricultural reasoning. A prominent advancement is the \textit{WisWheat} dataset introduced by \cite{yuan2025wiswheat}, which establishes a three-tiered vision-language benchmark encompassing image-caption pairs, visual question answering (VQA) for quantitative trait assessment, and conversational instruction tuning for field management. While \textit{WisWheat} significantly enhances the capacity of multimodal models to generate domain-specific textual recommendations, it treats textual descriptions sequentially. It lacks a structured symbolic repository to enforce structural integrity, eliminate linguistic noise, or anchor observations over prolonged temporal spans. To evaluate these vision architectures further, recent evaluation frameworks have pushed the boundaries of multi-task agronomic reasoning. The \textit{AgEval} benchmark developed by \cite{ageval2025} introduces a zero-shot and few-shot evaluation framework for plant stress phenotyping across twelve specialized subsets, including complex wheat rust classifications. Although multi-task benchmarks like \textit{AgEval} prove that foundation backbones can successfully handle stress identification and severity estimation, they process evaluation tasks dynamically as independent inputs. They lack a persistent, structured semantic architecture to store, link, and query those classification outputs across continuous long-term breeding cycles.

\item \textbf{Conversational Agents and Task Orchestration}
To move beyond isolated model inference, recent literature has explored multi-agent frameworks capable of executing complex phenotyping workflows. For instance, the \textit{PhenoAssistant} framework \cite{phenoassistant2026} orchestrates specialized computational agents via natural language interfaces to automate image preprocessing, feature extraction, and phenotypical data visualization. Although such multi-agent frameworks excel at procedural workflow execution and tool manipulation, they operate as functional pipelines rather than knowledge representation layers. They do not dynamically synthesize an integrated, persistent semantic network, nor do they enforce ontological consistency across disparate data acquisitions. This functional bottleneck has prompted the development of cross-modal sensing networks that layer language with multi-spectral features. The \textit{S3-Net} framework presented by \cite{s3net2025} implements an intelligent dual-spectral vision-language sensing network that links immediate visual features to broad botanical encyclopedic knowledge via a specialized Knowledge-Vision Alignment (KVA) module. While frameworks like \textit{S3-Net} leverage cross-modal semantic-guided discrimination to optimize few-shot variety and anomaly identification, their alignment mechanism is tailored purely to maximize static classification scores. They do not build or maintain an integrated, multi-hop knowledge graph structure to preserve these relationships over time.

\item \textbf{Textual Information Extraction and Entity Linking}
Extracting structured facts from static agricultural literature has traditionally relied on specialized corpora and Named Entity Recognition (NER) models. The \textit{Triticum aestivum Trait and Entity Corpus} (\textit{TaeC}), curated by \cite{nedellec2024taec}, provides a gold-standard manually annotated text dataset containing hundreds of PubMed abstracts. This corpus is optimized for training extraction engines to link textual variables directly to canonical biological ontologies like the Wheat Trait Ontology (\textit{WTO}). Nevertheless, these approaches remain strictly text-dominant and are engineered for peer-reviewed scientific literature. They lack the perception mechanisms required to handle the messy, unstructured languages of real-time agronomic field notes or map text terms back to localized image regions.

\item \textbf{Structural Graphs and Digital Twins}
Representing the physical layout of crops through graph topologies represents a robust mechanism for digital twin development. In \textit{CherryGraph}, developed by \cite{gilson2024cherrygraph}, the 3D topology of fruit tree canopies is programmatically encoded into a Resource Description Framework (RDF) knowledge graph database. This framework enables precise geometric queries regarding branch intersections and canopy structure. However, \textit{CherryGraph} relies exclusively on strict parametric sensor inputs and structural lidar configurations. It lacks the cross-modal flexibility required to ingest free-text multi-source data or align visual anomalies with biological state ontologies.
\end{itemize}

Concurrently, framework models like those introducing perceptual graph kernels \cite{fpls2026graphkernels} characterize crop health by modeling phenotypic variations as interacting multi-trait networks. Nodes within these networks represent distinct traits, while edges capture functional interdependencies and stress responses. Despite their modeling capacity, these graphs operate over abstract, pre-discretized feature labels. They do not leverage standardized public ontologies to normalize vocabulary across distinct experimental silos, nor do they support traceabilty back to the absolute bounding box or pixel segmentation masks of the raw crop images. To capture these underlying physical traits at a global scale, the domain has turned toward massive self-supervised architectures. The \textit{FoMo4Wheat} crop foundation model introduced by \cite{fomo4wheat2026} leverages millions of globally curated field images to achieve robust feature representation across diverse environmental conditions and canopy-level growth stages. However, because vision foundations like \textit{FoMo4Wheat} are purely vision-dominant, they lack an abstract symbolic or natural language interface, which prevents researchers from running semantic logical queries across the complex structures they interpret.
\subsection{Research Gaps}
\label{subsec:research_gaps}
Despite recent progress at the intersection of deep learning and precision agriculture, current methodologies suffer from three fundamental limitations:
\begin{enumerate}
    \item \textbf{The Cross-Modal Integration Gap:} Existing systems operate in isolated technical silos. State-of-the-art agronomic benchmarks focus heavily on text-based recommendations or question-answering systems, text mining frameworks extract localized entities from static peer-reviewed literature, and architectural digital twins map strict parametric 3D structural parameters. No unified architecture exists that can dynamically ingest unstructured, noisy field observations, reconcile them with standardized domain ontologies, and ground them directly onto multi-organ satellite or RGB canopy imagery.
    \item \textbf{Surface-Level Semantics:} Standard vision backbones and spatial-trait interaction models treat phenotypical features as independent variables or flat, sequential text blocks. Consequently, they are correlation-driven and lack explicit symbolic reasoning. Without an integrated neuro-symbolic layer capable of executing hierarchical class subsumption and multi-hop logical traversal, these models cannot support complex diagnostic querying or resolve vocabulary conflicts across disparate experiments.
    \item \textbf{The Traceability and Auditing Deficit:} Current end-to-end multi-agent orchestration pipelines or trait-graph engines optimize for procedural task execution or coarse macro-level assertions. They fail to preserve a mathematically traceable pixel-to-symbol audit trail. This makes it impossible to trace an abstract symbolic node in a graph (e.g., a specific fungal pathology) directly back to its exact spatial coordinates or bounding regions within the original high-resolution canopy image.
\end{enumerate}

\subsection{Core Contributions}
\label{subsec:contributions}
To systematically address these bottlenecks, this work proposes the \textit{PhenoNest} framework. The primary technical contributions are structured as follows:
\begin{itemize}
    \item \textbf{Separated Image Perception and Biological Logic:} We explicitly separate how the framework visually interprets crop images from how it reasons about biological facts. The pipeline utilizes a \textit{Perception Engine} (driven by a multimodal AI model) to generate detailed descriptions of plant traits and map out exact symptom locations, alongside a dedicated \textit{Logic Engine} designed to filter out conversational noise from field notes and extract clean biological relationships.
    \item \textbf{Centralized Snapshot Anchor (\texttt{Plant\_Obs\_Id}):} We introduce a unique, central tracking identifier, denoted as \texttt{Plant\_Obs\_Id}, which acts as a single temporal snapshot of a specific field observation. This anchor structurally links corresponding visual images and textual field notes, allowing researchers and breeders to track identical plant genotypes longitudinally over time, across different growth stages (phenophases), across distinct experimental plots, and under various environmental stressors.
    \item \textbf{Standardized Terminology Mapping and Concept Merging:} Using a specialized language model trained on botanical data (\textit{PlantDeBERTa}), the framework automatically translates informal or varied field notes into unique, universally recognized terms within standardized target databases, namely the Plant Ontology ($\mathcal{PO}$), Relationship Ontology ($\mathcal{RO}$), and Wheat Trait Ontology ($\mathcal{WTO}$). We implement an automated grouping layer that unites synonymous or overlapping naming variations of a trait under a single, unified definition while keeping each individual field observation completely distinct and traceable.
\end{itemize}

\section{Data Curation}

\subsection{Image Data}

\begin{figure}[H]
    \centering
    \includegraphics[width=0.55\linewidth]{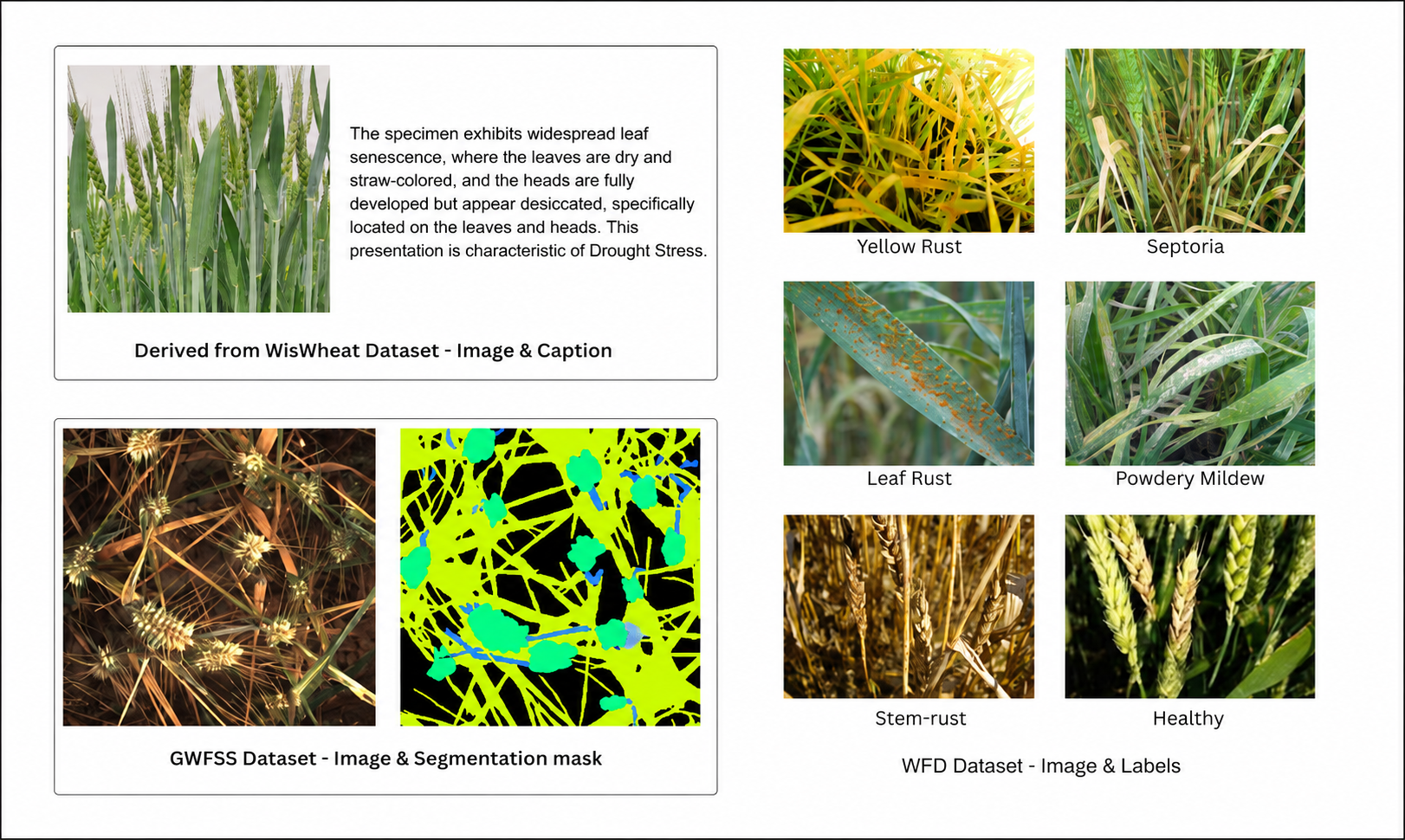}
    \caption{Visual comparison of the heterogeneous image datasets integrated into the framework, showcasing high-resolution field imagery from WisWheat, organ-level masks from GWFSS, localized bounding boxes from GWHD, pathological manifestations from WFD, and canopy segmentations from SegVeg.}
    \label{fig:dataset_visuals}
\end{figure}

The study integrates complementary visual datasets to provide organ-level spatial grounding, large-scale localization, and disease-induced variability.The structural properties and quantitative metrics of the data sets are compiled in Table~\ref{tab:dataset_characteristics}; additionally, Figure~\ref{fig:dataset_visuals} provides a comparative visual overview of the highly heterogeneous sample modalities and their respective annotations.

\begin{itemize}
    \item \textbf{WisWheat Dataset\cite{yuan2025wiswheat}:} Supplies raw high-resolution RGB field imagery and unstructured field notes.
    Within our framework, it serves as the primary multimodal anchor to test text distillation and validate generative spatial localization.

    \item \textbf{Global Wheat Field Semantic Segmentation (GWFSS) \cite{wang2025gwfss}:} Provides dense pixel-level masks for wheat heads, stems, and leaves.
    It is utilized as a spatial verification baseline to evaluate the model's accuracy in grounding localized traits onto specific plant organs.

    \item \textbf{Global Wheat Head Detection (GWHD)\cite{david2020global}:} Contains bounding-box annotations across diverse acquisition conditions.
    It serves as structural scaffolding to initialize coordinate anchors and guide instance-level visual grounding for wheat spikes.

    \item \textbf{Wheat Fungal Disease (WFD)\cite{genaev2021wheat}:} Features images exhibiting distinct fungal pathologies like rust and mildew.
    The framework leverages this dataset to benchmark the perception engine's capacity to classify fine-grained, disease-induced visual variations.

    \item \textbf{PlantPAD Dataset \cite{plantpad2024platform}:} Supplies rigid, parametric token tags and localized disease phenotype metadata. Within our framework, it serves as an evaluation baseline to benchmark the cross-dataset entity resolution pipeline and test the system's capacity to dense-populate unmapped morphological features directly from raw visual inputs.
\end{itemize}

\begin{table}[H]
\centering
\caption{Characteristics of datasets used in this study.}
\label{tab:dataset_characteristics}
\small
\begin{tabular}{l|c|c|c|c|c}
\hline
\textbf{Dataset} & \textbf{Task} & \textbf{Modality} & \textbf{\# Images} & \textbf{\# Samples} & \textbf{Annotation Type} \\ \hline
WisWheat & Instruction Tuning & RGB + Text & --      & 4,888              & Conversational           \\
GWHD     & Detection          & RGB        & 4,700   & 190,000 heads      & Bounding boxes           \\
GWFSS    & Segmentation       & RGB        & 53,174  & 1,096 labeled      & Pixel masks              \\

PlantPAD    & Classification       & RGB        & 421,314  & 22,132 labeled      & Phenotype Tags              \\
WFD2020  & Disease Classif.   & RGB        & 2,414   & 2,414              & Disease labels              \\ \hline
\end{tabular}
\end{table}

\subsection{Text Data}

The framework utilizes two primary textual resources to define phenotypic language and facilitate semantic extraction.

\begin{itemize}
    \item \textbf{WisWheat Field Notes\cite{yuan2025wiswheat}:} A collection of unstructured, noisy text data consisting of raw observations from agricultural trials.
    It provides the raw linguistic input parsed by our text distillation pipeline to isolate fine-grained phenotypes and environmental stressors from background agricultural noise.

    \item \textbf{Triticum aestivum Trait and Entity Corpus (TAEC)\cite{nedellec2024taec}:} A manually curated corpus of 540 PubMed articles annotated by domain experts for traits and phenotypes.
    It establishes the standardized biological language baseline to ensure that free-text annotations mirror peer-reviewed botanical literature rather than informal descriptions. It is leveraged for two primary tasks, i) Scientific Supervision - TAEC ensures that annotations follow established scientific phrasing and provides the necessary supervision.
    Operationally, it supplies the domain-expert labels required to train, fine-tune, and validate our named entity recognition (NER) models within the logic engine, and ii) Ontology Linkage - By mapping free-text descriptions to standardized trait concepts, TAEC enables ontology-aligned knowledge graph construction.
    It provides explicit semantic pathways to map extracted text terms directly into structured WTO, PO, and RO classes during graph integration.
\end{itemize}

\subsection{Ontologies}

The framework utilizes standardized biological ontologies to ensure extracted information is consistent and interpretable.

\begin{itemize}
    \item \textbf{Wheat Trait Ontology (WTO)\cite{nedellec2020wto},\cite{matteis2013crop}:} Represents wheat-specific traits and phenotype concepts at the organ and plant levels. The framework specifically maps concepts under the \textit{stress trait} branch (\texttt{CO\_321\_0000004}), tracking children nodes for \textit{biotic stress} (encompassing fungal, viral, and bacterial pathologies alongside animal damage) and \textit{abiotic stress} (including drought, waterlogging, and winter response anomalies).
    
    \item \textbf{Plant Ontology (PO)\cite{jaiswal2005po}:} Represents anatomical structures, such as wheat heads and stems, enabling consistent modeling of part-whole relationships. Operationally, our system integrates the entire \textit{plant structure} branch (\texttt{PO\_0009011}) to systematically map and resolve all physical plant organs and sub-components mentioned throughout the unstructured text data.
    
    \item \textbf{Plant Anatomy and Trait Ontology (PATO) \cite{pato2013attribute}:} Used to describe qualitative and quantitative attributes, including color, severity, or extent. Within this schema, the pipeline traverses the \textit{plant stress response trait} domain to extract the \textit{weed damage trait response} (\texttt{TO\_0000347}) sub-branch. It also leverages the \textit{whole plant morphology trait} (\texttt{TO\_0000398}) branch—specifically utilizing \textit{plant aspect} (\texttt{TO\_0000737}) and \textit{plant color} (\texttt{TO\_0000708})—to standardize macromorphological states.
\end{itemize}

\section{Methodology}

\subsection{Overview -- Framework Diagram}

\begin{figure}[H]
    \centering
    \includegraphics[width=0.6\textwidth]{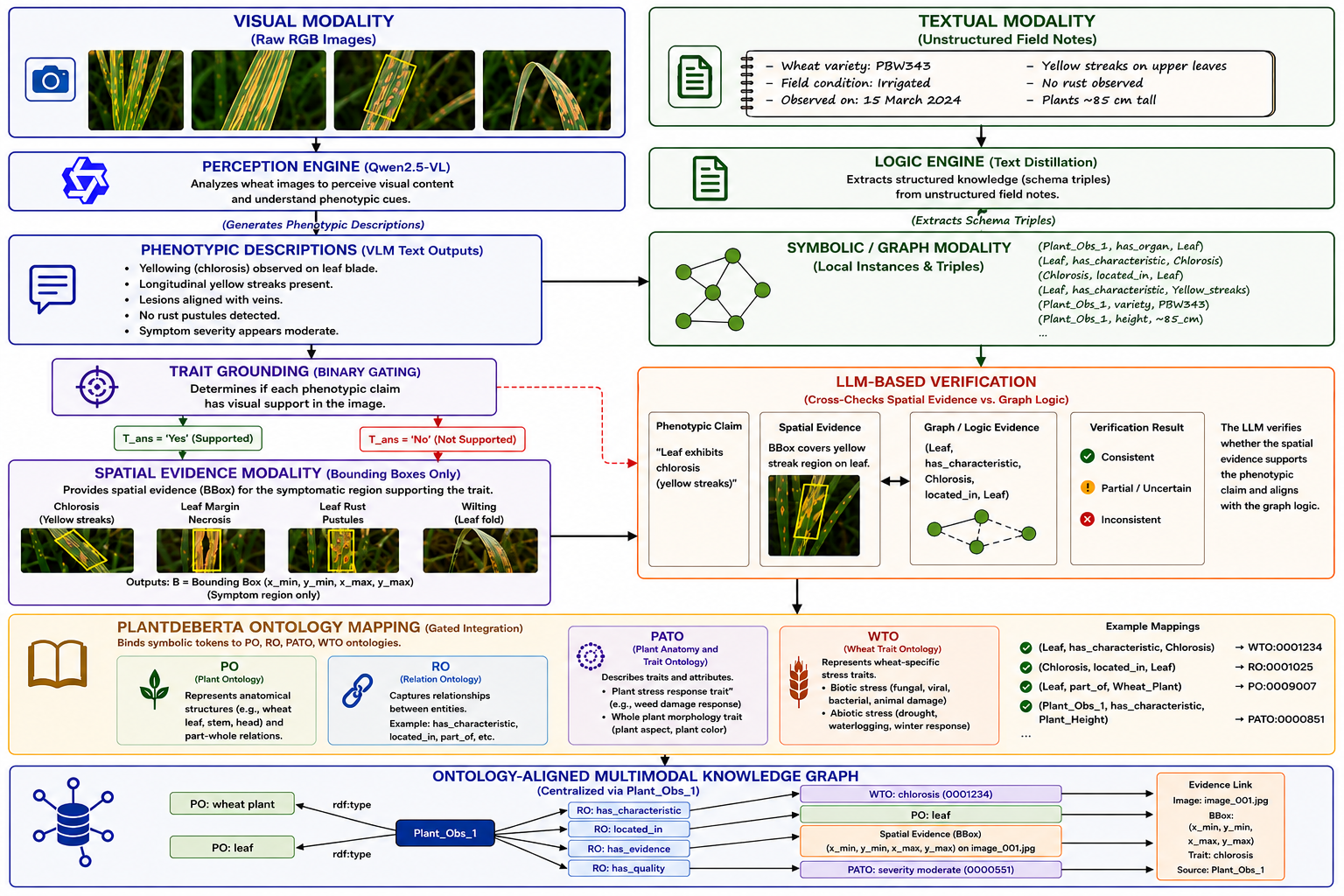}
    \caption{The pipeline integrates \textit{Visual} and \textit{Textual} modalities through a dual-engine extraction process, converging into a \textit{Spatial Evidence} and \textit{Ontology-Aligned Knowledge Graph} representation.}
    \label{fig:overall_architecture}
\end{figure}

\begin{figure}[H]
    \centering
    \includegraphics[width=0.65\textwidth]{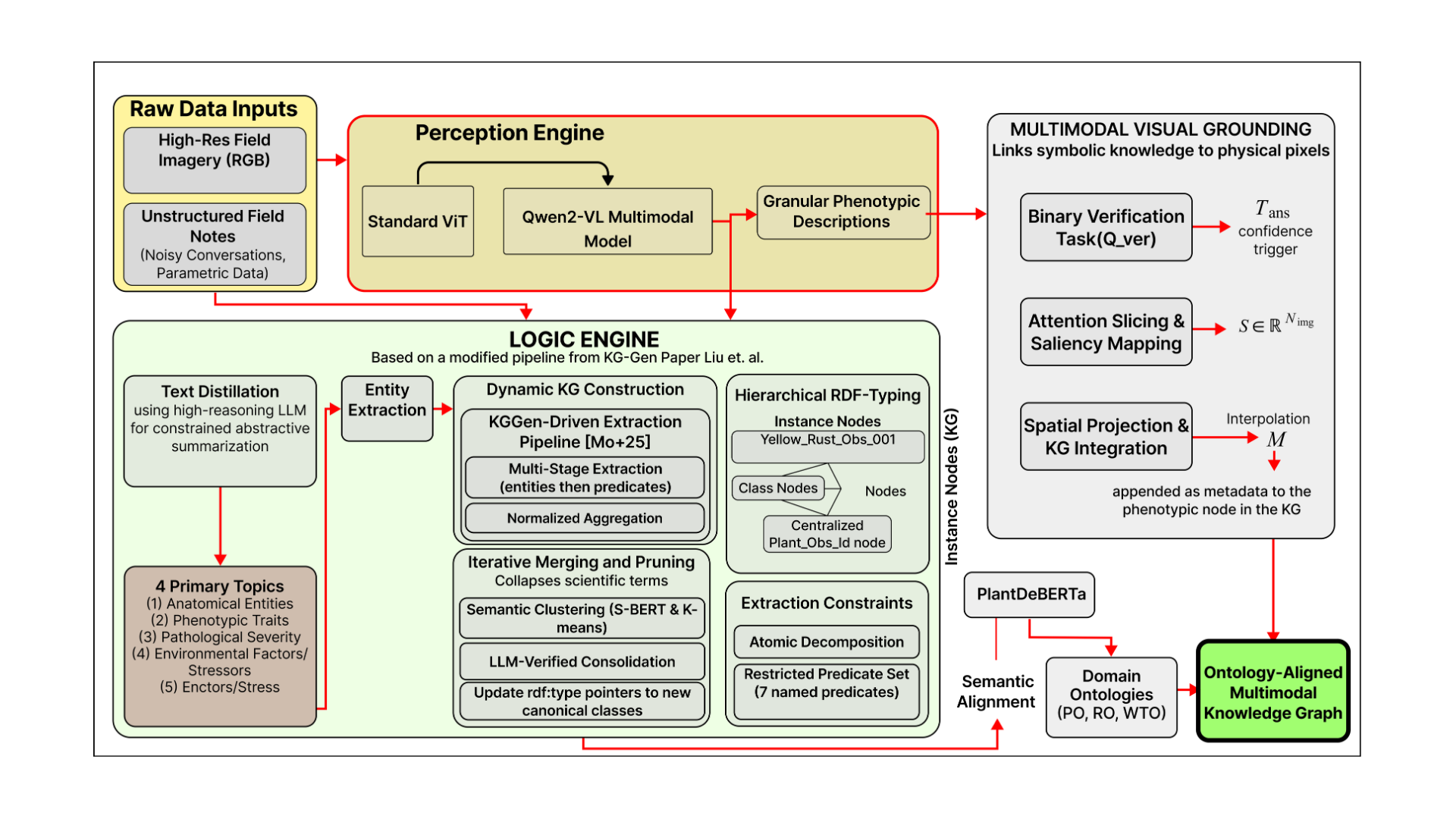}
    \caption{Detailed end-to-end architecture of the PhenoNest-MKG framework that builds the unified, queryable knowledge graph.}
    \label{fig:methodology_pipeline}
\end{figure}

This section details the proposed overall PhenoNest framework flow as shown in Figure~\ref{fig:overall_architecture}. The methodology is structured around a novel two-engine pipeline that explicitly separates visual perception from semantic reasoning to ensure both modularity and precision in high-throughput phenotyping, the explicit end-to-end architecture for this pipeline is detailed in Figure~\ref{fig:methodology_pipeline}. The first component, the Perception Engine, leverages the Qwen2-VL multimodal model\cite{Qwen2VL}.

The second component, the Logic Engine, processes this extracted textual data through a dedicated Large Language Model (LLM) pipeline, this is a modified pipeline built upon work done by Mo et~al.\ \cite{mo2025kggen}. This engine filters noise and dynamically constructs a Multimodal Knowledge Graph (MKG) anchored by temporal observation nodes, that represent a ``snapshot'' of the plant at that instance. Through RDF-typing and contextual embeddings generated by PlantDeBERTa, these nodes are semantically aligned with standardized domain ontologies (PO, RO, and WTO). Finally, the framework executes spatial visual grounding, physically linking the VLM-generated attention maps to the formally structured graph nodes to enable traceable, cross-modal phenotypic reasoning.

\subsection{Text Distillation and Entity Extraction}

The raw textual data from the WisWheat dataset comprises unstructured conversations generated by human annotators about an image; these conversations combine morphology and phenotypic traits with operational metadata (e.g., irrigation logs or nutrient management). To maintain the semantic purity of the PhenoNest, we implemented a pre-processing distillation pipeline designed to decouple biological signals from agricultural noise.

This phase leverages a high-reasoning Large Language Model (LLM) to perform constrained abstractive summarization. The model is prompted to act as a domain-specific filter, isolating segments referring exclusively to plant morphology, growth stages, and pathological symptoms. This summarization method leverages the studies done by Liu et~al.\ \cite{liu2024learning}, where such summaries from large models are used to train smaller models. These distilled text summaries are used as annotations/captions for the images from the dataset for training VLMs for phenotype extraction, similar to a weakly supervised learning approach similar to Zhang et. al. \cite{fpls2025weaklyvlm}.

The distillation is structured such that the refined text contains information about four primary topics:
\textbf{(i) Anatomical Entities:} identifying localized organs (e.g., spikes, flag leaves); \textbf{(ii) Phenotypic Traits:} capturing morphological states (e.g., chlorosis, senescence); \textbf{(iii) Pathological Severity:} normalizing qualitative and quantitative stress indicators; and \textbf{(iv) Environmental Factors/Stressors:} abiotic or biotic stressors affecting growth (e.g., Drought Stress, Moisture Deficiency, Potassium Deficiency, Heat Stress).

\subsection{Dynamic KG Construction}
The construction of the granular MKG, PhenoNest serves as the framework's primary reasoning layer, where unstructured phenotypic observations are transformed into a formal, queryable knowledge base. Unlike static knowledge graphs, this construction process is designed to handle the high-velocity, multimodal data inherent in plant breeding. By integrating the perception results from the Vision-Language Model (VLM) with the logic results from distilled field notes, the framework creates a unified temporal record of plant health.The sequential logic for this initial triple extraction and entity consolidation is formalized in Algorithm~\ref{alg:dynamic_kg_construction}.

\subsubsection{The KGGen-Driven Extraction Pipeline}

To establish the foundational entities and relationships within the graph, the framework adopts the KGGen pipeline \cite{mo2025kggen}. This method is specifically chosen to mitigate the ``sparsity'' problem---where a graph contains nearly as many unique relation types as it does edges by enforcing structural constraints during the initial extraction phase. \textbf{(i) Multi-Stage Extraction:} Rather than an end-to-end extraction, the system employs a two-step process: first identifying a comprehensive list of entities (subjects and objects) and subsequently determining the specific predicates (relations) that connect them. \textbf{(ii) Normalized Aggregation:} Extracted triples are collected across multiple sources and normalized to a canonical lowercase format to reduce initial redundancy before the graph is fully instantiated. \textbf{(iii) Logical Consistency:} By using DSPy-based structured outputs \cite{khattab2024dspy}, the Large Language Model (LLM) is constrained to follow strict instructions, ensuring that every subject-predicate-object triple is faithful to the original source text. Operationally, this logical consistency is enforced by injecting explicit prompt-based constraint logic into the high-reasoning extraction layer, forcing atomic decomposition, noun normalization, and strict predicate restriction. This prompt logic maps unstructured text sequences into deterministic symbolic representations as demonstrated in Table~\ref{tab:extraction_constraints_example_1}.

\begin{table}[H]
\centering
\caption{In-Context Prompt Logic Constraints for Multi-Modal Triple Extraction}
\label{tab:extraction_constraints_example_1}
\small
\begin{tabular}{p{3.5cm}|p{3.5cm}|p{5.5cm}}
\hline
\textbf{Prompt Constraint Rule} & \textbf{Linguistic Operation} & \textbf{DL Framework Prompt Logic / Output Triples} \\ \hline
\textbf{Atomic Decomposition} & Deconstruct complex phrases into individual biological units. & \textit{Raw:} ``severe yellow-orange pustules'' \newline $\rightarrow$ \texttt{(pustule, has\_color, yellow-orange)} \\ \hline
\textbf{Noun Normalization} & Convert descriptive adjectives and verbs into canonical nouns. & \textit{Raw:} ``The leaves are shriveled.'' \newline $\rightarrow$ \texttt{(leaf, exhibits\_symptom, shriveling)} \\ \hline
\textbf{Intensifier Removal} & Strip subjective qualifiers to ensure objective graph density. & \textit{Raw:} ``highly prevalent tiny spots'' \newline $\rightarrow$ \texttt{(spot, rdf:type, instance)} \\ \hline
\textbf{Restricted Predicate Set} & Enforce strict alignment to the 7 biological predicates. & \textit{Allowed Predicates:} \newline \texttt{exhibits\_symptom, caused\_by, is\_at\_stage, has\_color, has\_part, associates\_with\_trait, has\_state} \\ \hline
\end{tabular}
\end{table}

\subsubsection{Hierarchical RDF-Typing and Instance Generation}

Following the identification of entity classes via the KGGen logic, the framework applies a custom RDF \cite{w3crdf2014} Typing layer to facilitate longitudinal monitoring.An overview of this stage can be viewed in the  This allows the system to distinguish between a general biological concept (a ``Class'') and a specific, time-stamped observation (an ``Instance''). \textbf{(i) Instance Instantiation:} For every identified phenotype in an image or field note, a unique instance node is generated (e.g., \textit{rdf:type:Yellow\_Rust\_Obs\_001}). \textbf{(ii) Semantic Anchoring:} These instances are linked to a centralized \textit{Plant\_Obs\_Id} node, which stores the temporal metadata and experiment IDs necessary for tracking the same plant across different phenophases. \textbf{(iii) Dynamic Class Assignment:} Every instance is assigned to its respective class node via an \textit{rdf:type} relation. If a required class (e.g., ``Chlorosis'') is not already present in the existing graph, the system automatically generates a new class node to maintain structural integrity.

\subsubsection{Iterative Merging and Relational Pruning}

To ensure the graph remains concise and scientifically accurate as the experiment progresses, an iterative merging phase is executed. This phase collapses synonymous scientific terms while preserving the independence of individual observations in three steps. \textbf{(i) Semantic Clustering:} The system utilizes S-BERT \cite{reimers2019sentencebert} to generate embeddings for all Class Nodes, which are then grouped using a K-means clustering \cite{macqueen1967kmeans}. \textbf{(ii) LLM-Verified Consolidation:} A high-reasoning LLM reviews these clusters to identify synonymous concepts, such as merging ``Stripe Rust'' and ``Yellow Rust''. The LLM selects the most scientifically accurate canonical alias to serve as the unified class node. \textbf{(iii) Relational Preservation:} During this pruning phase, the unique Instance Nodes remain distinct; however, their \textit{rdf:type} pointers are updated to point to the new canonical class. This architecture ensures that the graph ``collapses'' at the conceptual level for easier querying while remaining ``expanded'' at the observation level for precise trait tracking.

\subsubsection{Extraction Constraints and Semantic Integrity}

To ensure that the MKG remains scientifically rigorous and computationally dense, the framework applies a series of high-level constraints during the KGGen-driven extraction phase. These constraints are enforced through specialized prompting strategies governing entity and relationship identification.

\begin{itemize}
\item \textbf{Entity Extraction Constraints}

The extraction process is governed by several linguistic and biological rules designed to create a standardized data structure: (i) \textbf{Atomic Decomposition}: complex observations are deconstructed into atomic units, separating the Plant Organ from its Morphological State (e.g., “Pustule” and “Yellow-orange” instead of “yellow-orange pustules”); (ii) \textbf{Noun Normalization}: adjectives and verbs are converted into canonical nouns (e.g., “Shriveled” to “Shriveling”); and (iii) \textbf{Intensifier Removal}: subjective qualifiers such as “severe” or “slight” are omitted to ensure objective data entries.These constraints map the messy free-text into atomic entities. as detailed in Table~\ref{tab:extraction_constraints_example}.
\begin{table}[H]
\centering
\caption{Instruction-Constrained Triple Extraction and Normalization Mapping}
\label{tab:extraction_constraints_example}
\small
\begin{tabular}{p{4.5cm}|p{3.2cm}|l|p{3.2cm}}
\hline
\textbf{Raw Input Field Note} & \textbf{Subject (Entity)} & \textbf{Predicate} & \textbf{Object (Value/Entity)} \\ \hline
\multirow{4}{=}{\textit{``The flag leaves are showing severe chlorosis, and we found tiny brown rust pustules on the lower canopy stems.''}} & \texttt{flag\_leaf} & \texttt{exhibits\_symptom} & \texttt{chlorosis} \\
 & \texttt{stem} & \texttt{has\_part} & \texttt{lower\_canopy} \\
 & \texttt{lower\_canopy} & \texttt{exhibits\_symptom} & \texttt{pustule} \\
 & \texttt{pustule} & \texttt{has\_color} & \texttt{brown} \\ \hline
\end{tabular}
\end{table}

\item \textbf{Relationship Schema and Directionality}

Relationship extraction is restricted to a strictly defined set of predicates to maintain graph density:
\textbf{(i) Restricted Predicate Set:} the framework is limited to seven biological predicates: \textit{exhibits\_symptom}, \textit{caused\_by}, \textit{is\_at\_stage}, \textit{has\_color}, \textit{has\_part}, \textit{associates\_with\_trait}, and \textit{has\_state}; and \textbf{(ii) Mandatory Health Connection:} a triple defining the overall health status (e.g., (Wheat $|$ has\_state $|$ Diseased)) must be included for every record.
\end{itemize}

\begin{algorithm}[H]
\small
\caption{Dynamic Knowledge Graph Construction and Merging}
\label{alg:dynamic_kg_construction}
\begin{algorithmic}[1]
\REQUIRE Text inputs $\{T_{\text{field}}, P_{\text{vlm}}\}$, valid predicates $\mathcal{P}$, model $M_{\text{sbert}}$, clusters $K$.
\ENSURE Local Graph $\mathcal{G}_{\text{local}}$ anchored to $\text{Plant\_Obs\_Id}$.
\STATE $\mathcal{G}_{\text{local}} \leftarrow \{(\text{Plant\_Obs\_Id}, \text{rdf:type}, \text{ex:Observation})\}$
\STATE $\mathcal{T}_{\text{cand}} \leftarrow \text{KGGen}\left(\text{FilterConstraints}(T_{\text{field}} \cup P_{\text{vlm}})\right)$
\FOR {each $(s, p, o) \in \mathcal{T}_{\text{cand}}$ \textbf{where} $p \in \mathcal{P}$}
    \STATE $I_o \leftarrow \text{UniqueInstance}(\text{lowercase}(o))$
    \STATE $\mathcal{G}_{\text{local}} \leftarrow \mathcal{G}_{\text{local}} \cup \{(I_o, \text{ex:prop\_of}, \text{Plant\_Obs\_Id}), (I_o, \text{rdf:type}, C_o), (\text{lowercase}(s), p, I_o)\}$
\ENDFOR
\STATE $\mathcal{U}_{1 \dots K} \leftarrow \text{KMeans}(\text{M}_{\text{sbert}}(\text{GetClasses}(\mathcal{G}_{\text{local}})), K)$
\FOR {each cluster $\mathcal{U}_i$ \textbf{where} $\mathcal{S} \leftarrow \text{LLM\_Synonyms}(\mathcal{U}_i) \neq \emptyset$}
    \STATE $C_{\text{canonical}} \leftarrow \text{LLM\_SelectName}(\mathcal{S})$
    \FOR {each $C_{\text{var}} \in \mathcal{S} \setminus \{C_{\text{canonical}}\}$}
        \STATE $\text{Redirect}(C_{\text{var}} \rightarrow C_{\text{canonical}})$
        \STATE $\mathcal{G}_{\text{local}} \leftarrow \mathcal{G}_{\text{local}} \setminus \{C_{\text{var}}\}$
    \ENDFOR
\ENDFOR
\STATE \textbf{assert} $(\text{Wheat}, \text{has\_state}, \text{Diseased}) \in \mathcal{G}_{\text{local}}$
\RETURN $\mathcal{G}_{\text{local}}$
\end{algorithmic}
\end{algorithm}

\subsection{Semantic Alignment and Knowledge Graph Integration}

The final layer of the PhenoNest-MKG framework governs semantic alignment, bridging the localized, generative representations produced by the Multimodal Visual Grounding block with formal domain ontologies (PO, RO, WTO). This phase ensures that the continuous coordinates and extracted text are unified into a mathematically consistent, queryable neuro-symbolic space.The end-to-end mapping sequence and verification rules are formalized in Algorithm~\ref{alg:semantic_alignment_integration}.

\subsubsection{Ontology Mapping via Domain-Specific Language Models}

To reconcile natural language variations across field records, the framework utilizes \textbf{PlantDeBERTa} \cite{plantdeberta2025}, a transformer model specialized for botanical nomenclature.
\textbf{(i) Input Processing:} the model ingests the canonical classes stabilized during iterative merging alongside raw textual entities from the text distillation phase; and \textbf{(ii) Hierarchical Assignment:} PlantDeBERTa projects these terms into a high-dimensional vector space, mapping them via cosine similarity thresholds to unique concept identifiers within three core target schemas: the Plant Ontology (PO) for anatomy, the Relationship Ontology (RO) for predicates, and the Wheat Trait Ontology (WTO) for disease phenotypes.

\subsubsection{LLM-Based Verification and Graph Integration}

Before the knowledge graph is permanently updated, the compiled multi-modal outputs undergo an \textbf{LLM-Based Verification} step to enforce structural and factual consistency across the visual perception and textual logic streams. The consistency is checked via a two part sequence:
\textbf{(i) Consistency Check:} The Logic Engine ingests the candidate triples extracted from the field notes alongside the descriptive phenotype text generated by the VLM Perception Engine. The LLM reviews these parallel streams to ensure there are no biological or contextual contradictions. \textbf{(ii) Gated Integration:} This verification acts as a strict binary gate. If the VLM's visual detection tags and the extracted field note text are symbolically aligned, the transaction clears. Only at this stage does the framework transition to data persistence: the continuous visual coordinate bounding sets $\mathcal{B}$ and the reference URI pointing to the discrete pixel hardmask matrix $\mathcal{H}$ are compiled into a standardized, structured JSON block. This multi-modal metadata payload is then appended directly as an explicit data property field to the unique instance node. Conversely, if any topological or cross-modal contradiction is flagged, the entire database transaction is aborted immediately to safeguard graph integrity.By validating the extracted text against the VLM's perceptual findings before committing the spatial metadata ensures that every symbolic relationship written into the graph is backed by matching visual evidence, preserving an auditable path from abstract database concepts directly to the physical image assets.
\begin{algorithm}[H]
\caption{Semantic Alignment and Spatial Multimodal Integration}
\label{alg:semantic_alignment_integration}
\begin{algorithmic}[1]
\REQUIRE Local Graph $\mathcal{G}_{\text{local}}$, Ontologies $\mathcal{O}$, $\text{PlantDeBERTa}$, Threshold $\tau$, Layout $\mathcal{B}$, Mask $\mathcal{H}$.
\ENSURE Aligned Multimodal Knowledge Graph $\mathcal{G}_{\text{final}}$.
\STATE $\mathcal{G}_{\text{final}} \leftarrow \mathcal{G}_{\text{local}}$
\FOR {each $c \in \text{GetClassNodes}(\mathcal{G}_{\text{final}})$}
    \STATE $\omega^* \leftarrow \arg\max_{\omega \in \mathcal{O}} \left( \text{CosineSim}(\text{PlantDeBERTa}(c), \text{PlantDeBERTa}(\omega)) \right)$
    \IF {$\text{CosineSim}(\text{PlantDeBERTa}(c), \text{PlantDeBERTa}(\omega^*)) \ge \tau$}
        \STATE $\mathcal{G}_{\text{final}} \leftarrow \mathcal{G}_{\text{final}} \cup \{(c, \text{owl:equivalentClass}, \text{GetURI}(\omega^*))\}$
    \ENDIF
\ENDFOR
\FOR {each Instance Node $I_x \in \mathcal{G}_{\text{final}}$ \textbf{with} visual bounds}
    \STATE $V_{\text{lbl}} \leftarrow \text{GetVLMDescription}(I_x)$
    \IF {$\text{LLM\_Verify}(\text{Type}(I_x), V_{\text{lbl}}) = \text{``Consistent''}$}
        \STATE $\mathcal{G}_{\text{final}} \leftarrow \mathcal{G}_{\text{final}} \cup \{(I_x, \text{ex:has\_spatial\_evidence}, \text{JSON}(\mathcal{B}, \mathcal{H}))\}$
    \ELSE
        \STATE \textbf{abort transaction};\ $\text{LogAnomaly}(I_x)$
    \ENDIF
\ENDFOR
\RETURN $\mathcal{G}_{\text{final}}$
\end{algorithmic}
\end{algorithm}
\subsection{Multimodal Visual Grounding}

Multimodal Visual Grounding serves as the critical intersection in the PhenoNest-MKG framework, where high-dimensional visual features extracted by the Perception Engine are physically and semantically mapped to structured entities within the KG. While traditional vision-language models often rely on implicit, unnormalized global correlations or noisy internal attention layers prone to diffusion across deep transformer blocks, this stage provides explicit spatial anchors for phenotypic reasoning. By shifting from heuristic feature slicing to generative visual grounding, the framework directly links localized pixel coordinates to biological traits and phenotypic states, ensuring every symbolic representation within the graph is explicitly anchored to physical data. The grounding process utilizes the natively aligned multi-dimensional coordinate projections of the Qwen2.5-VL engine. These coordinates are then structured into visual metadata payloads and linked to the instance nodes associated with a specific field experiment created during the dynamic KG construction phase. This explicit mapping ensures that each node in the PhenoNest-MKG is not merely a symbolic label but is structurally grounded in localized visual evidence, ensuring every biological claim is supported and traceable.

\subsubsection{Binary Verification Task}

For each phenotypic relation $R(e_i, t_j)$ extracted, where $e_i$ is an anatomical entity (e.g., leaf, stem) and $t_j$ is a phenotypic trait or anomaly (e.g., red brown pustules), the system formulates a closed-ended verification query $Q_{ver}$. The prompt structure forces a dual-constraint response requiring an initial binary validation followed by explicit localization instructions. The model's primary sequence generation acts as a confidence trigger: if the affirmative token $T_{ans} = \text{`Yes'}$ is emitted, the spatial localization pipeline initializes; if $T_{ans} = \text{`No'}$, the relation is flagged as unverified, and the visual grounding subroutine for that specific instance is aborted.

\subsubsection{Generative Spatial Localization and Graph Integration}
To provide explicit visual evidence for verified phenotypic traits without introducing interpolation artifacts, the framework leverages the Perception Engine (Vision Language Model eg. QwenVL-2) to provide discrete, normalized bounding box sequences. Let $I_{\text{field}}$ represent the high-resolution RGB input from the WisWheat dataset. For an activated confidence trigger, the decoder parses the verification query $Q_{\text{ver}}$ to produce a collection $\mathcal{B}$ of $K$ target bounding regions:
\begin{equation}
    \mathcal{B} = \left\{ b_k \;\middle|\; b_k = [y_{\text{min}}, x_{\text{min}}, y_{\text{max}}, x_{\text{max}}]_k, \, k \in \{1, 2, \dots, K\} \right\}
\end{equation}
To structurally anchor these boundaries within the \textit{PhenoNest-MKG}, a deterministic parser maps the coordinate sets $\mathcal{B}$ and their associated semantic labels to the graph environment. For tasks requiring regional localization, a discrete hardmask matrix $\mathcal{H} \in \{0, 255\}^{\text{height} \times \text{width}}$ is instantiated as a zero-matrix and populated for areas where the bounding box is present.
The raw coordinate list $\mathcal{B}$, label metadata, and the reference URI to the mask array $\mathcal{H}$ are compiled into a standardized JSON payload. This payload is appended directly to the corresponding phenotypic node, establishing a verifiable visual audit trail that connects abstract symbolic knowledge to the exact physical pixels of the crop canopy.
\section{Experimental Set Up}

\subsection{VLM Performance Metrics}

The framework is evaluated across three primary dimensions: generative descriptive quality, semantic alignment, and spatial grounding accuracy.

\subsubsection{Generative Quality (LLM-as-a-Judge)}

To evaluate the descriptive accuracy of the Qwen2-VL engine, a decomposed LLM-as-a-Judge framework \cite{zheng2023judging} is employed. Rather than evaluating a comprehensive output in a single pass, which often introduces cognitive bias and oversight, the framework utilizes an \textbf{item-by-item verification strategy}. The judge is prompted to evaluate exactly one phenotypic entity at a time against the VLM description. This granular approach ensures that the judge focuses on semantic alignment for specific traits without being overwhelmed by long lists, leading to higher reliability and more precise ablation data.

To formalize these Generative Quality Metrics without relying on standard token-matching n-gram overlap scores (such as BLEU or ROUGE \cite{papineni2002bleu, lin2004rouge}) which fail to capture biological synonymy, we pass the unstructured text descriptions into a batch in-context LLM-as-a-Judge parsing layer. Let $\mathcal{C}_i = \{c_{i,1}, c_{i,2}, \dots, c_{i,N}\}$ be the set of reference phenotypic checklist items for sample $i$, and let $\hat{\mathcal{V}}_i = \{v_{i,1}, v_{i,2}, \dots, v_{i,N}\}$ be the binary verdict vector generated by the evaluator model, where $v_{i,j} \in \{0, 1\}$ represents the semantic presence or absence of item $c_{i,j}$ within the VLM output description.

The Global Recall ($R_{\text{global}}$) is defined as the total proportion of verified phenotypic facts recovered across the entire evaluation dataset:
\begin{equation}
R_{\text{global}} = \frac{\sum_{i=1}^{M} \sum_{j=1}^{|C_i|} v_{i,j}}{\sum_{i=1}^{M} |C_i|}
\end{equation}
where $M$ denotes the total number of samples. To safeguard the ablation study against model verbosity and hallucination tendencies, a proxy Precision ($P_{\text{proxy}}$) metric is established by introducing an independent negative control checklist item $c_{\text{neg}}$ (representing a plausible but completely absent agricultural anomaly) to each sample description:
\begin{equation}
P_{\text{proxy}} = \frac{\sum_{i=1}^{M} (1 - v_{i,\text{neg}})}{M}
\end{equation}
The Macro F1-Score ($F_1$) is subsequently computed as the harmonic mean of the Global Recall and the proxy Precision to yield a unified index of generative performance:
\begin{equation}
F_1 = 2 \cdot \frac{P_{\text{proxy}} \cdot R_{\text{global}}}{P_{\text{proxy}} + R_{\text{global}}}
\end{equation}

We establish two distinct sub-dimensions of recall within this setup: Observational Recall ($R_{\text{obs}}$) specifically measures the recovery of immediate physical, anatomical, or macromorphological crop traits (e.g., leaf shape, stem structure), while the Inferred Recall ($R_{\text{inf}}$) isolates higher-level diagnostic or situational assertions (e.g., cultivation challenges, ecosystem parameters). For each required trait within its respective domain, the judge outputs a binary value $V \in \{\text{Yes}, \text{No}\}$.

\subsection{Visual Word Sense Disambiguation (VWSD) for Perception Engine Evaluation}

Visual Word Sense Disambiguation (VWSD) evaluates a Vision-Language Model's (VLM) ability to resolve linguistic ambiguity using localized visual features. In high-throughput plant phenotyping, raw descriptive terms are highly ambiguous due to overlapping visual manifestations of different crop diseases or weed presence. Standard text-matching models fail to ground these features accurately. To benchmark the visual grounding limits of open-source baselines ($\text{Qwen2.5-VL-3B}$ and $\text{Qwen2.5-VL-7B}$ \cite{bai2025qwen3vl}), we implemented a sequential binary classification experiment.

Rather than passing multiple candidate images simultaneously which hazards Out-of-Memory (OOM) VRAM exceptions due to visual token grid explosions—the framework streams evaluation images into the target network sequentially ($1$-by-$1$). For a single evaluation challenge, the model is presented with one True Positive image and four Distractor images representing distinct phenotypic anomalies. At each streaming step, the model parses the individual image against a randomly selected ambiguous text context template. The system records the model's internal log-probabilities (logits) assigned to the positive target token (\texttt{YES}). A challenge is marked as a successful hit if and only if the absolute maximum probability mass across the $5$-image deck uniquely tracks to the True Positive sample. This probabilistic selection eliminates hard-string affirmation bias while maintaining strict OOM safety.

The benchmark isolates five specific agronomic and botanical categories. To prevent the models from optimizing for static semantic formulations, we engineered a randomized context pool consisting of three distinct descriptive variants per target class. The structural setup and specific textual templates used during inference are outlined in Table~\ref{tab:vwsd_setup}.

\begin{table}[H]
\centering
\caption{VWSD Categorical Setup and Randomized Context Matrix}
\label{tab:vwsd_setup}
\small
\begin{tabular}{l|p{4cm}|p{6.5cm}}
\hline
\textbf{Category} & \textbf{Reference Target Condition} & \textbf{Sample Ambiguous Context Template} \\ \hline
\textbf{Healthy} & Uniform canopy with optimal vegetative development. & \textit{``The foliar landscape shows optimal chlorophyll accumulation with completely smooth margins and clean leaf surfaces.''} \\ \hline
\textbf{Leaf Rust} & Scattered pustules erupting through the blade epidermis. & \textit{``Foliar surfaces display small, isolated orange-brown powdery pustules erupting unevenly across the blade epidermis.''} \\ \hline
\textbf{Septoria} & Elongated necrotic lesions containing pycnidial speckling. & \textit{``Foliar structures show advanced speckled necrotic blotches, with dark elongated dead zones eating into the blade.''} \\ \hline
\textbf{Powdery Mildew} & Superficial white ash-like mycelial configurations. & \textit{``The leaf surface is covered in fuzzy, ash-white web-like configurations spreading over the dorsal canopy.''} \\ \hline
\textbf{Weed Presence} & Non-graminaceous broadleaf canopy interference. & \textit{``The crop rows reveal early broadleaf interference, with young lobed weed elements breaking through the soil surface.''} \\ \hline
\end{tabular}
\end{table}

\begin{itemize}
    \item \textbf{Sequential Binary Inference:} Loading images individually forces the VLM to compute visual attention weights strictly over the local pixel matrix of a single crop or leaf. It avoids token length scaling limits ($\mathcal{O}(N^2)$ attention complexity) and establishes an absolute, clear evaluation baseline.
    \item \textbf{Logit-Based Probabilistic Selection:} General-purpose VLMs exhibit a severe text-generation affirmation bias, frequently outputting hard text choices like \texttt{YES} for multiple related inputs. Extracting the raw token log-probabilities and executing a relative $\max()$ function forces a strict choice, surfacing the model's true internal preference.
    \item \textbf{Randomized Multi-Context Pool:} Deploying three distinct descriptions per category changes the textual embedding vectors passed to the model across different runs. This validates whether the model is truly matching visual symptoms to botanical concepts, or simply over-fitting to specific token arrangements in the prompt.
\end{itemize}

\subsection{Case Studies Setup}

The utility of the PhenoNest-MKG framework is evaluated through three targeted case studies designed to demonstrate the transition from unstructured multimodal observations to an ontology-aligned, queryable knowledge base. These scenarios address the core challenges of data fragmentation, semantic ambiguity, and longitudinal monitoring in wheat phenotyping. To ensure rigorous consistency across all evaluations, a single, unified KG was deployed globally across every case study. This underlying graph was constructed from an evaluation dataset of 150 samples curated from the WisWheat and PlantPAD \cite{plantpad2024platform} datasets, explicitly selected to cover a representative spectrum of wheat diseases and pathological symptoms. The evaluation suite is structured into the following targeted case studies:

\subsubsection{Setup 1: Cross-Dataset Entity Resolution Pipeline}
To demonstrate the framework's capacity to neutralize data silos, this study compares phenotypic observations across disparate acquisition environments. Because PlantPAD relies on flat, rigid parametric tokens, generating a graph solely from its text metadata results in a highly sparse structure with critical biological information gaps \cite{bioner2026lossmasking}. To mitigate this data-silo limitation, the framework leverages its Image Phenotype Extractor (VLM Phase) directly on the specimen's visual features. While the text metadata remains sparse, the visual pipeline aggressively extracts unmapped morphological anomalies and micro-phenotypes from the image, significantly enriching the PlantPAD graph. 

Because the source graphs are constructed in complete isolation, their initial structural iterations differ significantly due to vocabulary and tokenization discrepancies. To resolve these differences, the framework executes its \textbf{Semantic Alignment Engine}, which performs a two-stage operational routine:
\begin{enumerate}
    \item \textbf{Linguistic Base-Meaning Reduction:} Prior to formal ontology matching, extracted entities are programmatically resolved down to their root biological definitions. Token variations are lemmatized and normalized so that morphological and text-density shifts are compressed.
    \item \textbf{Ontological Class Mapping:} This reduction is applied directly to the generated Class URIs across both subgraphs. The normalized local concepts are then mapped to standard domain ontologies (WTO, PO), explicitly bridging the gap between syntactically distinct but biologically identical nodes.
\end{enumerate}

\subsubsection{Setup 2: Ontology-Subsumption Driven Traversal Rules}
This scenario utilizes the hierarchical architecture of the Wheat Trait Ontology (WTO) coupled with the DeepOnto toolkit \cite{he2024deeponto} to facilitate deep conceptual reasoning and multi-level semantic discovery across isolated observation records. The operational rules governing this traversal space are structured as follows:
\begin{itemize}
    \item \textbf{Concept-Aware Search Strategy:} When a user executes a natural language search query, a Large Language Model (LLM) first processes the input to identify the necessary target phenotype classes residing within the local \textit{ex:} namespace. Because the Semantic Alignment Engine has already bound these custom runtime classes to their canonical ontology counterparts (e.g., mapping \textit{ex:Wheat\_Leaf\_Rust} to WTO's Leaf Rust concept), the framework gains immediate access to the standard domain taxonomy.
    \item \textbf{Subsumption and Inverted Traversal Mechanism:} To surface structurally similar or pathologically related records, the framework utilizes DeepOnto to parse the ontology structure, compute subsumption properties, and identify shared parent nodes. The engine walks up the taxonomy to locate the lowest common ancestor (LCA) node such as the \textit{Fungal Disease} parent branch. From this high-level abstraction, the engine explores sibling branches to discover hidden or related disease classes (e.g., discovering Septoria Blotch as a sibling fungal pathology):
    \begin{equation}
        \text{Leaf Rust} \sqsubseteq \text{Fungal Disease} \quad \sqcap \quad \text{Septoria Blotch} \sqsubseteq \text{Fungal Disease}
    \end{equation}
    \item \textbf{Instance Shortcutting via Property Anchors:} Once a related sibling ontology class is discovered, the framework maps it back down to its corresponding local \textit{ex:} class namespace. The execution layer isolates the specific individual instances linked to this class via standard \textit{rdf:type} predicates. Finally, because every extracted phenotypic node in the PhenoNest-MKG maintains a direct structural anchor back to its root observation node via the custom \textit{ex:property\_of} predicate, the framework traverses this single-hop shortcut to immediately return the complete matching plant observation record (\textit{Plant\_Obs\_X}) to the user.
\end{itemize}

\subsubsection{Setup 3: Automated Downstream Dataset Generation Pipeline}
To evaluate the operational autonomy and accuracy of the framework, this scenario demonstrates how the PhenoNest-MKG acts as an end-to-end, automated dataset curation engine for downstream machine learning tasks, completely eliminating the need for manual graph engineering. The algorithmic pipeline maps raw data streams through the following steps:
\begin{itemize}
    \item \textbf{Zero-Overhead Automated Construction:} The data ingestion phase is fully automated. The user is only required to supply the raw specimen images along with their shallow, accompanying text metadata or parametric tags. The framework’s Image Phenotype Extractor (VLM phase) automatically parses the visual traits, handles local entity extraction, and builds the foundational knowledge graph structures dynamically behind the scenes.
    \item \textbf{Trait-Driven Retrieval and Semantic Fallback:} To curate a targeted downstream training set, a user supplies an unconstrained textual query describing desired phenotypic traits. The framework parses this query and attempts a two-tier resolution: (1) mapping extracted query tokens to canonical ontology classes (e.g., \textit{Leaf Rust}, \textit{Mildew}), and (2) capturing highly descriptive, non-ontological entities by computing vector-space semantic similarity against the unmapped local attributes extracted during the VLM phase.
    \item \textbf{Image Harvesting via Property Anchors:} Once instances satisfying the target trait intersections are isolated, the framework traverses the single-hop \texttt{ex:property\_of} predicate back to the root observation node (\texttt{Plant\_Obs\_X}). The system automatically harvests the registered source image paths, compiling a clean, partitioned dataset ready for downstream model training.
\end{itemize}
\subsection{UI Development: Interactive Interface and System Architecture}

\begin{figure*}[htbp!]
    \centering
    \begin{subfigure}[b]{0.49\textwidth}
        \centering
        \includegraphics[width=\textwidth]{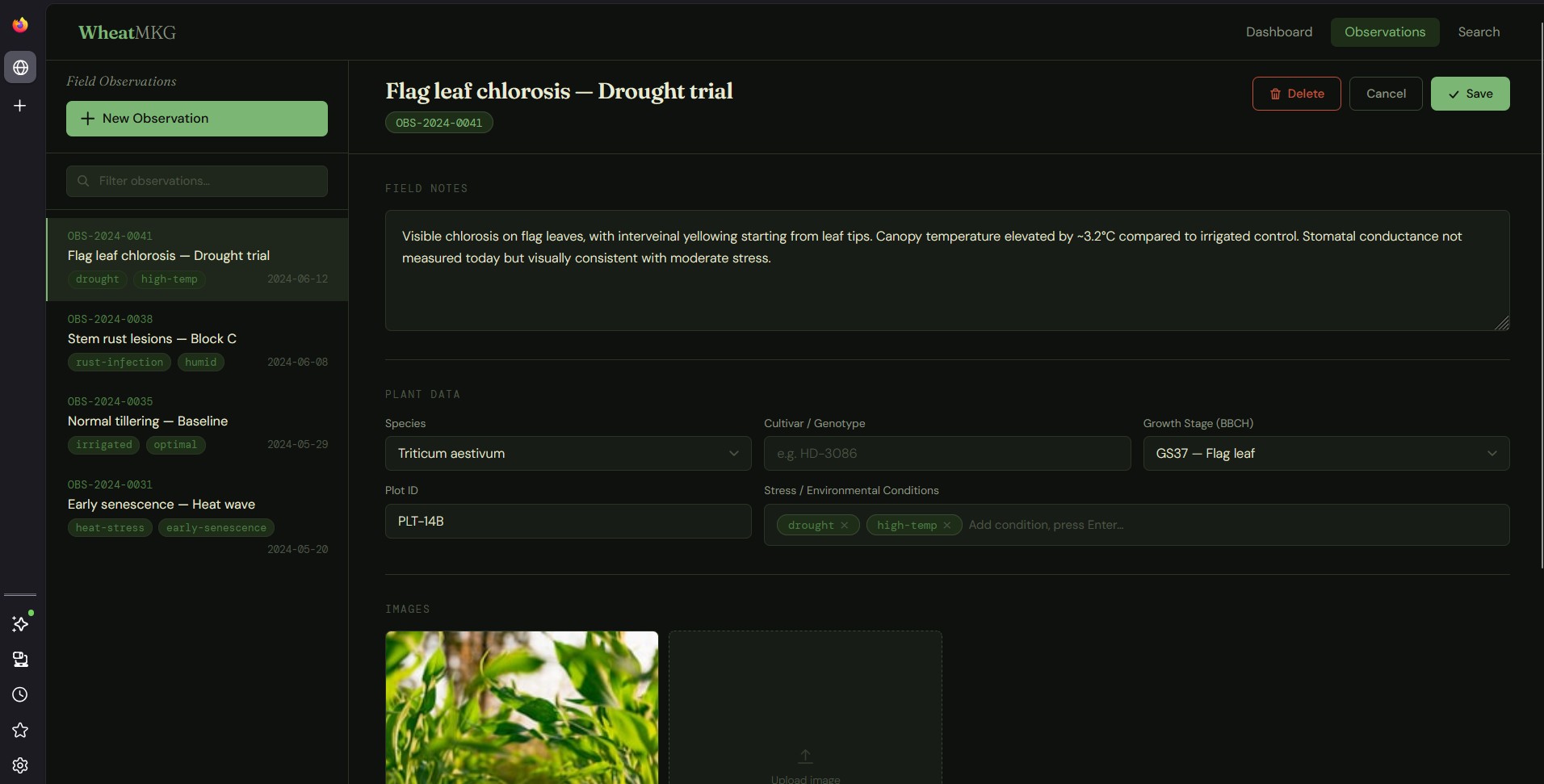}
        \caption{Screen 2: Master-detail observation manager panel tracking multi-modal fields and core ontology instances.}
        \label{fig:ui_screen2}
    \end{subfigure}
    \hfill
    \begin{subfigure}[b]{0.49\textwidth}
        \centering
        \includegraphics[width=\textwidth]{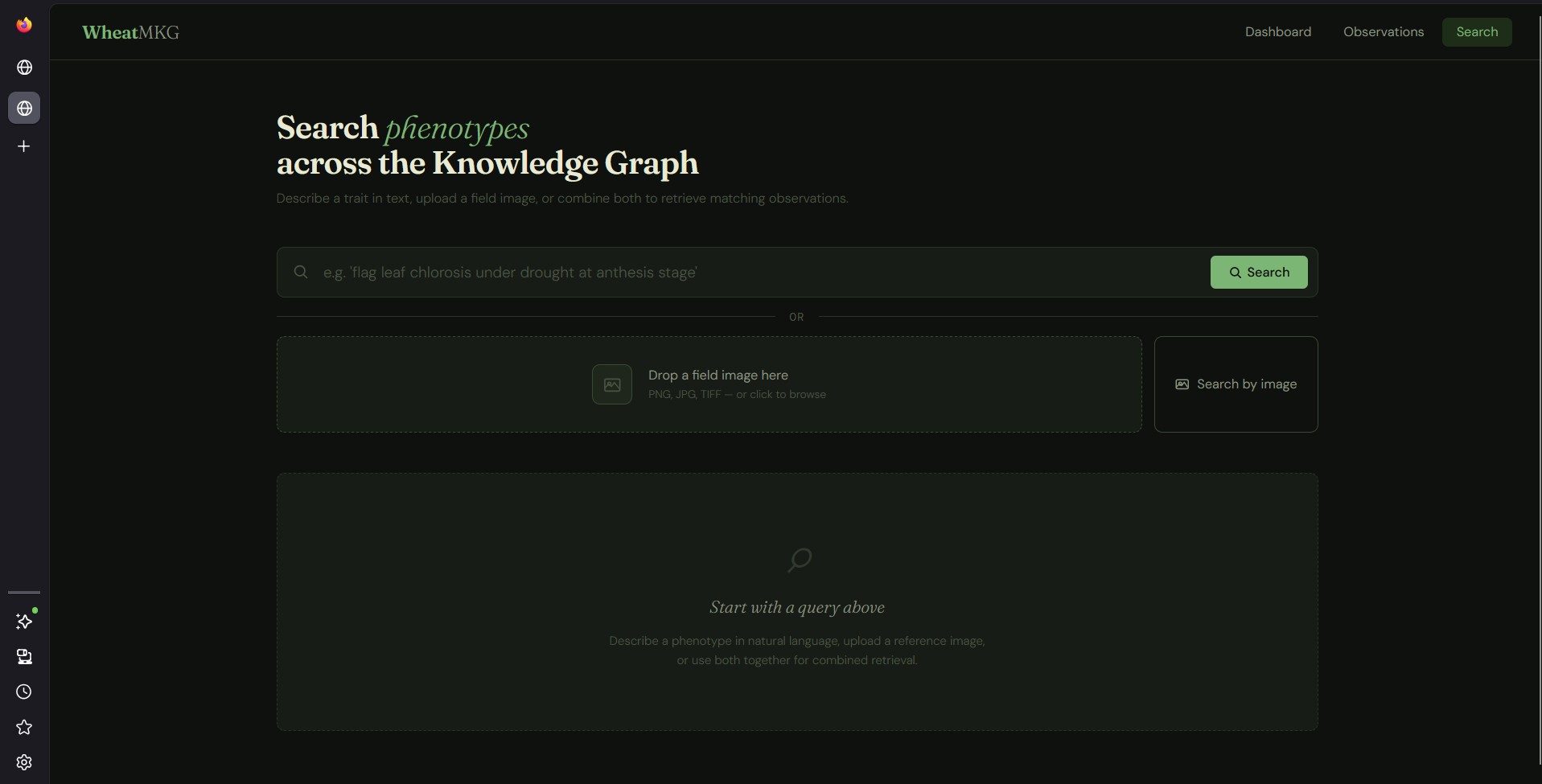}
        \caption{Screen 3: Unified natural language and visual query interface returning grounded match provenance badges.}
        \label{fig:ui_screen3}
    \end{subfigure}
    \caption{Interactive web interface of the PhenoNest-MGKG framework displaying (a) the data curation dashboard and (b) the multimodal search layout}
    \label{fig:framework_ui_dashboards}
\end{figure*}

To bridge the gap between our underlying multi-modal knowledge graph, PhenoNest and practical field deployment, we developed a lightweight, high-performance web interface implemented via a Flask  backend \cite{ronacher2010flask} and a custom semantic frontend. The user interface is intentionally structured into two core operational screen, as shown in Figure~\ref{fig:framework_ui_dashboards}, to facilitate real-time agricultural data curation and intuitive discovery:

\begin{itemize}
    \item \textbf{Observation Manager (Screen 1):} This screen acts as the primary data curation hub using a master-detail dashboard layout. The left sidebar lists all historical observations, filterable by name or ID, while the right panel displays a comprehensive input form for the selected record. Researchers use this form to log unstructured field notes, biological ontologies (species and cultivar), BBCH crop growth stages, plot configurations, variable micro-phenotype bounding regions ($\mathcal{B}$), and temporal links. Once submitted, this data is compiled into a structured metadata string and ingested directly into the Knowledge Graph as properties of a centralized observation node (\texttt{Plant\_Obs\_Id}).
    
    \item \textbf{Multimodal Search Interface (Screen 2):} This screen provides a unified interface for dual-modality graph querying. It features a natural language text search bar and an asynchronous, drag-and-drop image upload zone, which can be used independently or together for hybrid search. Upon execution, the frontend sends these multi-source payloads via asynchronous JavaScript \texttt{fetch} requests to the backend logic engine. The system processes the input against the knowledge base and returns prioritized result cards marked with a matching confidence score, highlighted text snippets, and metadata tags indicating the match provenance (\textit{TEXT}, \textit{IMAGE}, or \textit{TEXT+IMAGE}).
\end{itemize}

\section{Results and Discussion}

\subsection{Performance Metrics and Illustrations}

To evaluate the semantic parsing and visual grounding accuracy of the ablated vision-language baselines, we execute the evaluation matrix across our structural validation sets. The empirical findings compiled in Table~\ref{tab:ablation_matrix} illustrate a clear scaling performance gap between the $3\text{B}$ and $7\text{B}$ parameter network tiers. While both models demonstrate an acceptable capacity for handling broader inferred traits, their scores drop sharply when evaluating fine-grained spatial and taxonomic markers.

\begin{table}[H]
\centering
\caption{Multimodal Phenotypic Ablation Evaluation Matrix}
\label{tab:ablation_matrix}
\begin{tabular}{l|c|c|c|c}
\hline
\textbf{VLM Baseline} & \textbf{Macro F1} & \textbf{Global Recall} & \textbf{Obs.\ Recall} & \textbf{Inf.\ Recall} \\ \hline
Qwen-VL-3B            & 0.1918            & 10.6\%                 & 11.6\%                & 6.7\%                 \\
Qwen-VL-7B            & 0.2631            & 15.1\%                 & 16.1\%                & 11.4\%                \\ \hline
\end{tabular}
\end{table}

To diagnose the precise nature of these failures, the evaluations are categorized into context-specific behavioral categories in Table~\ref{tab:deep_behavioral_insights}.

\begin{table}[H]
\centering
\caption{Deep Behavioral Insights Performance Matrix}
\label{tab:deep_behavioral_insights}
\small
\begin{tabular}{l|c|c|c|c}
\hline
\textbf{VLM Baseline} & \textbf{Spatial Density} & \textbf{Spatial Location} & \textbf{Negative Assertions} & \textbf{Taxonomic Identity} \\ \hline
Qwen-VL-3B & 4.6\% (16/345)  & 13.2\% (105/795) & 55.9\% (57/102) & 5.9\% (26/438)  \\
Qwen-VL-7B & 9.0\% (31/345)  & 16.0\% (127/795) & 63.7\% (65/102) & 11.0\% (48/438) \\ \hline
\end{tabular}
\end{table}

As shown in Table~\ref{tab:deep_behavioral_insights}, both baseline models acceptably perform on \textit{Negative Assertions} ($55.9\%$ for $3\text{B}$ and $63.7\%$ for $7\text{B}$). This indicates that recognizing clean, healthy crop states and confirming the absence of disease is relatively straightforward for general-purpose vision backbones, as healthy canopies match highly prevalent text distributions in common training datasets. However, performance plummets when confronting \textit{Spatial Density} ($4.6\%$) and \textit{Taxonomic Identity} ($5.9\%$) at the $3\text{B}$ scale. This structural limitation highlights a fundamental failure in low-parameter vision encoders, which struggle to preserve the pixel resolution necessary to ground tiny, localized weed seedlings scattered against complex soil backgrounds.

Scaling the model parameter footprint to $7\text{B}$ nearly doubles the extraction rate for both density parameters ($9.0\%$) and exact scientific taxonomic matching ($11.0\%$). \textit{Spatial Location} also yields an improvement from $13.2\%$ to $16.0\%$, validating that fine-grained agricultural trait grounding scales predictably with parameter volume. Despite these improvements, the low absolute performance values across spatial and taxonomic categories highlight the necessity of an independent reasoning infrastructure such as a Semantic Alignment Engine to resolve gaps left by the perception layer.


\begin{table}[H]
\centering
\caption{Granular Model Transformation Blueprint and Categorized Extractions}
\label{tab:data_snapshot}
\small
\begin{tabular}{p{4.5cm}|p{4.5cm}|c|c}
\hline
\textbf{VLM Raw Unstructured Output Snippet} & \textbf{Target Reference Checklist Item} & \textbf{Category} & \textbf{Score} \\ \hline
\multirow{4}{4.5cm}{\raggedright \textit{``A wild radish (\textit{Raphanus raphanistrum}) weed is seen growing alongside the wheat. It shows characteristic lobed leaves but the density is quite low on the soil surface.''}}
 & Wild radish classified as a broadleaf annual weed. & Taxonomic Identity & 1 \\ \cline{2-4}
 & Wild radish plants characterized by lobed leaves. & Spatial Location & 1 \\ \cline{2-4}
 & Low-density presence of wild radish. & Spatial Density & 1 \\ \cline{2-4}
 & Severe structural rust damage is present. & Negative Control & 0 \\ \hline
\end{tabular}
\end{table}

Table~\ref{tab:data_snapshot} presents a concrete data snapshot highlighting how unstructured model text maps directly into categorized binary evaluation metrics. Moving from shallow, direct question-answering systems to this structured checklist parsing model allows the framework to extract maximum actionable metadata from raw text streams.
\subsection{Visual Word Sense Disambiguation (VWSD) Evaluation}

The empirical results of the sequential binary Visual Word Sense Disambiguation (VWSD) task in Table~\ref{tab:vwsd_results} are presented below. The benchmark evaluates the capacity of both models to successfully isolate a single true agronomic profile out of a five-image deck populated with high-resemblance visual distractors.

\begin{table}[H]
\centering
\caption{Visual Word Sense Disambiguation (VWSD) Multi-Scale Performance Matrix}
\label{tab:vwsd_results}
\small
\begin{tabular}{l|c|c|c|c|c}
\hline
\textbf{VLM Baseline} & \textbf{Overall Accuracy} & \textbf{Healthy} & \textbf{Leaf Rust} & \textbf{Septoria} & \textbf{Powdery Mildew} \\ \hline
Qwen-VL-3B            & 44.0\% (11/25)            & 80.0\% (4/5)     & 20.0\% (1/5)       & 20.0\% (1/5)      & 20.0\% (1/5)            \\ 
Qwen-VL-7B            & 64.0\% (16/25)            & 100.0\% (5/5)    & 40.0\% (2/5)       & 60.0\% (3/5)      & 40.0\% (2/5)            \\ \hline
\end{tabular}
\end{table}

The metrics highlight distinct behavioral patterns and boundaries within the models' visual classification systems. At the lower tier, the $\text{Qwen-VL-3B}$ baseline achieves an overall strict challenge accuracy of $44.0\%$. While this clears the $20.0\%$ baseline of random guessing, the breakdown reveals that the model's performance is heavily skewed by structural categories. The model displays high accuracy on clear classification targets, hitting an $80.0\%$ success rate on both \textit{Healthy} tissue and \textit{Weed Presence of Radish}. However, it drops to a minimal $20.0\%$ accuracy on complex pathogenic conditions like \textit{Leaf Rust}, \textit{Septoria}, and \textit{Powdery Mildew}.

The primary error profiles explain the root cause of these systemic failures. The $3\text{B}$ parameter encoder frequently confuses \textit{Leaf Rust} with \textit{Septoria}, and misidentifies \textit{Powdery Mildew} as \textit{Septoria} in four out of five challenges. Because these diseases share overlapping visual signatures such as chlorotic discoloration, brown spotting, and structural tissue degradation, the smaller model lacks the resolution capacity to distinguish fine leaf variations in a vacuum. It exhibits an over-grounding bias, reacting generally to any visible spot or lesion rather than separating distinct fungal features.

Scaling the perception architecture to the $\text{Qwen-VL-7B}$ baseline yields an overall strict accuracy improvement to $64.0\%$. The larger parameter footprint resolves general ambiguities completely, pushing the \textit{Healthy} verification category to a perfect $100.0\%$ success rate. Significant performance gains are also recorded across challenging diagnostic symptoms, with \textit{Septoria} identification accuracy rising to $60.0\%$. This improvement confirms that expanding the vision encoder's capacity allows the model to process subtle visual details, such as distinguishing tiny pycnidial black speckles from scattered rust pustules. 

Despite these architectural improvements, the $7\text{B}$ model still encounters systematic limits, capping at a $40.0\%$ success rate for \textit{Leaf Rust} and \textit{Powdery Mildew}. These persistent visual errors emphasize that raw vision-language perception layers remain vulnerable to fine-grained feature confusion when evaluated on sequential data streams. This finding directly supports the core hypothesis of this framework: a standalone VLM is insufficient for reliable high-throughput plant phenotyping.

\subsection{Case Studies: Validation and Empirical Results}

\subsubsection{Cross-Dataset Entity Resolution Outcomes}
\begin{figure}[H]
    \centering
    \includegraphics[width=0.5\textwidth]{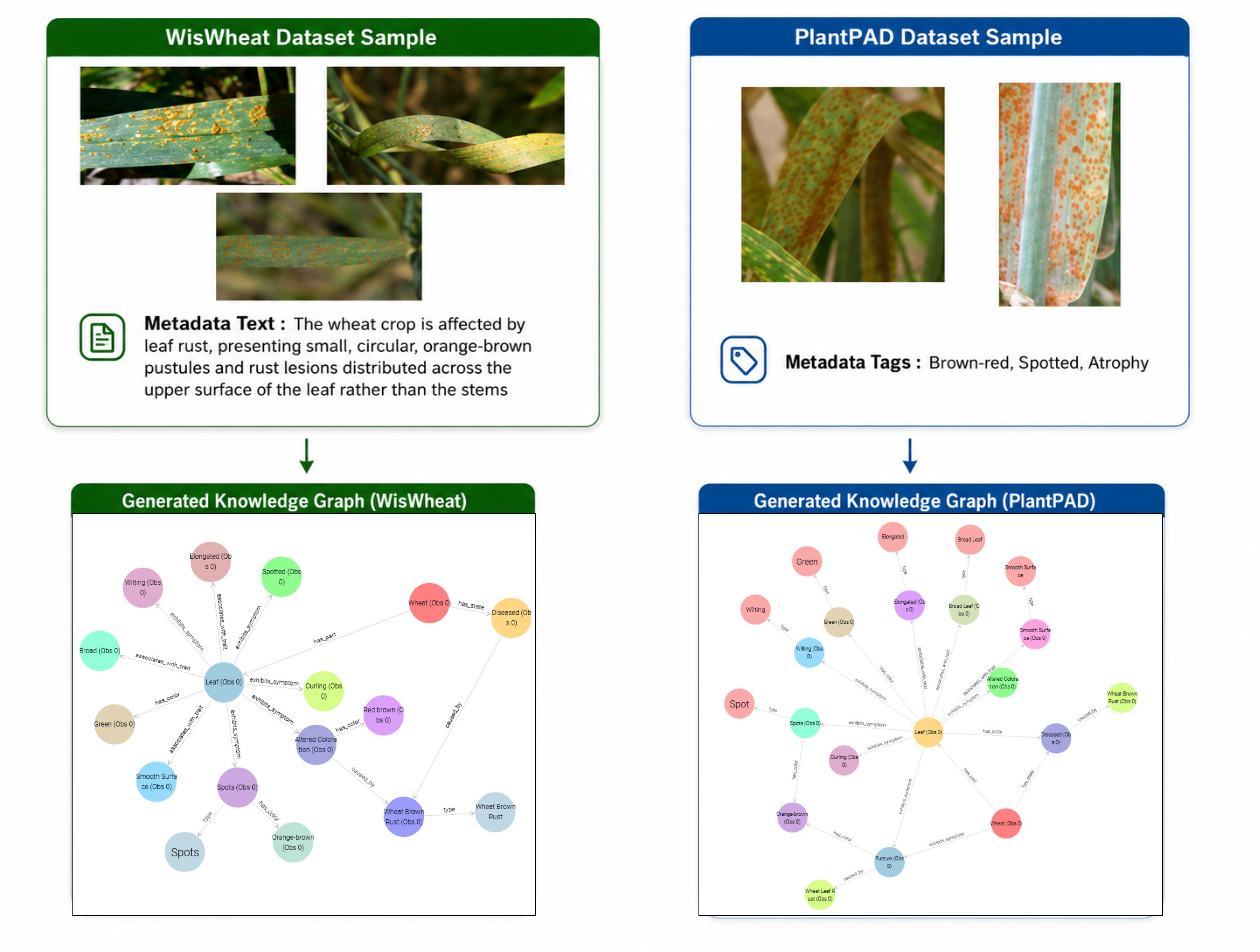}
    \caption{Visual architecture of Case Study 1 (Cross-Dataset Entity Resolution). The upper row contrasts the heterogeneous inputs from distinct source silos: a text-dense, unstructured sample from the WisWheat dataset and a token-sparse parametric sample from the PlantPAD dataset.}
    \label{fig:cross_dataset_resolution}
\end{figure}

The empirical deployment of the pipeline successfully demonstrated the integration of fragmented data environments. As illustrated in Figure~\ref{fig:cross_dataset_resolution}, the framework processed highly disparate inputs: rich, text-heavy unstructured samples from the WisWheat dataset and parametric, token-sparse tags from the PlantPAD dataset. 

Building knowledge graphs purely from standalone, token-sparse metadata tags, such as those in the PlantPAD dataset-leads to substantial data omissions because rigid parametric labels cannot capture the granular details of a physical specimen. By executing the Image Phenotype Extractor directly on the raw visual targets, the framework successfully recovers these unmapped phenotypes, dynamically populating the graph with critical biological information.

Furthermore, because these source graphs are constructed in complete isolation, they inherently suffer from naming conflicts due to vocabulary discrepancies across independent datasets. The Semantic Alignment Engine successfully neutralizes these conflicts through a two-stage operational routine. It first performs a linguistic base-meaning reduction by lemmatizing and normalizing token variations to compress text-density shifts. It then maps these stabilized custom runtime concepts directly to standardized canonical domain schemas, such as the Wheat Trait Ontology ($\mathcal{WTO}$) and Plant Ontology ($\mathcal{PO}$). This dual-stage alignment effectively bridges the gap between syntactically distinct but biologically identical nodes, establishing absolute semantic equivalence across the distinct dataset namespaces. This evaluation paradigm directly mirrors the established framework used in large-scale entity matching profiling benchmarks, such as the comprehensive study by Primpeli and Bizer \cite{primpeli2020profiling}, which systematically analyzes entity resolution systems using this exact triad of metrics to determine structural correctness. Furthermore, our metric configuration aligns with domain-specific cross-resource entity linking evaluations, such as the framework implemented by Kartchner et al. \cite{kartchner2023comprehensive}.

\subsubsection{Ontology-Subsumption Driven Reasoning Results}
 When broad conceptual lookups were executed, the framework successfully navigated inherited properties within the target ontologies to harvest hidden or unlinked related records. The quantitative metrics evaluated for this exploratory discovery track are summarized in Table~\ref{tab:subsumption_discovery_metrics}.  The design of our hierarchical routing evaluation and the profiling of structural traversal metrics (such as topological hops and path distance profiles) match established methodologies in neuro-symbolic knowledge engineering. To evaluate the efficiency of our taxonomic traversal routines, we align our benchmarking setup with the platform standards introduced by He et al. \cite{he2024deeponto} in the development of \textit{DeepOnto}, which formalizes the use of automated reasoners to extract inferred subsumers, compute taxonomic paths.As analyzed in structural ontology evaluation frameworks like those by Elmke et al. \cite{elmke2025experts}, strict hierarchical constraints and rigid top-level schema partitioning frequently introduce downstream reasoning limitations, where the lack of shared inter-domain ancestors or lower-level Lowest Common Ancestor (LCA) hubs breaks semantic continuity and isolates concurrent phenotypic features during multi-hop graph querying.
\begin{table}[H]
\centering
\caption{Ontological Subsumption Exploration and Boundary Metrics}
\label{tab:subsumption_discovery_metrics}

\begin{subtable}{\textwidth}
\centering
\caption{Intra-Domain Routing (Biotic Branch Traversal Validation)}
\label{tab:subsumption_success}
\begin{tabular}{lcccc}
\toprule
\textbf{Discovered Sibling Class} & \textbf{WTO Identity} & \textbf{LCA Parent Link} & \textbf{Curation Status} & \textbf{Harvested Yield} \\
\midrule
Powdery Mildew & \texttt{CO\_321\_0000114} & \text{fungal disease} & Verified Match & 19 Observations \\
Rust           & \texttt{CO\_321\_0000133} & \text{fungal disease} & Verified Match & 18 Observations \\
Blotch         & \texttt{CO\_321\_0000134} & \text{fungal disease} & Data Absent    & 0 Observations  \\
\bottomrule
\end{tabular}
\vspace{0.3cm}
\end{subtable}

\begin{subtable}{\textwidth}
\centering
\caption{Cross-Domain Routing Boundary Case (Biotic-Abiotic Structural Disconnection)}
\label{tab:subsumption_failure}
\begin{tabular}{lcccc}
\toprule
\textbf{Target Sibling Class} & \textbf{WTO Identity} & \textbf{LCA Parent Link} & \textbf{Curation Status} & \textbf{Harvested Yield} \\
\midrule
Drought Stress & \texttt{CO\_321\_0000034} & \text{abiotic stress} & \text{Verified Match}   & 15 Observations \\
Leaf Rust      & \texttt{CO\_321\_0000133} & \text{fungal disease} & \text{Failed Routing} & 0 Observations  \\
Powdery Mildew & \texttt{CO\_321\_0000114} & \text{fungal disease} & \text{Failed Routing} & 0 Observations  \\
\bottomrule
\end{tabular}
\vspace{0.3cm}
\end{subtable}

\begin{tabular}{lclc}
\multicolumn{4}{l}{\textbf{Exploratory Routing Capability Metrics:}} \\
\midrule
$\bullet$ Discovered Taxonomic Sibling Branches: & 3 & $\bullet$ Verification Path Latency: & 14 ms \\
$\bullet$ Shortest Topological Path: & 6 hops & $\bullet$ Cross-Domain Path Break: & \textbf{Infinite Hops} \\
\bottomrule
\end{tabular}
\end{table}
The empirical findings compiled in Table~\ref{tab:subsumption_discovery_metrics} validate both the efficiency of our ontology-subsumption routing mechanisms and the critical structural boundaries imposed by existing schema taxonomies. As documented in Table~\ref{tab:subsumption_success}, when performing intra-domain queries confined to the biotic stress branch, the logic engine functions seamlessly. By passing local nodes through DeepOnto, the framework successfully identified 3 distinct taxonomic sibling branches under the shared global Lowest Common Ancestor ($\text{LCA}$) parent hub (\texttt{CO\_321\_0000122: fungal disease}). This traversal achieved a path latency of just $14\text{ ms}$ over 6 topological hops, realizing a harvested yield of 19 observations for \textit{Powdery Mildew} and 18 observations for \textit{Rust}. A third sibling branch (\textit{Blotch}) was successfully mapped but yielded 0 observations due to real-world data gaps in the raw evaluation profiles.

However, a major structural limitation was uncovered when executing searches on multi-stressor samples containing concurrent biotic and abiotic symptoms. As evaluated in the boundary experiment in Table~\ref{tab:subsumption_failure}, when the system processes an initial target token representing an abiotic condition such as \textit{Drought Stress} (\texttt{CO\_321\_0000034}), the query paths correctly to 15 matching records. Yet, even though these exact sample instances concurrently present active \textit{Leaf Rust} and \textit{Powdery Mildew} visual symptoms on the leaf blade, the reasoning engine returns a harvested yield of 0 observations for both diseases, triggering a ``Failed Routing'' status. 

\begin{figure}[H]
    \centering
    \includegraphics[width=\linewidth]{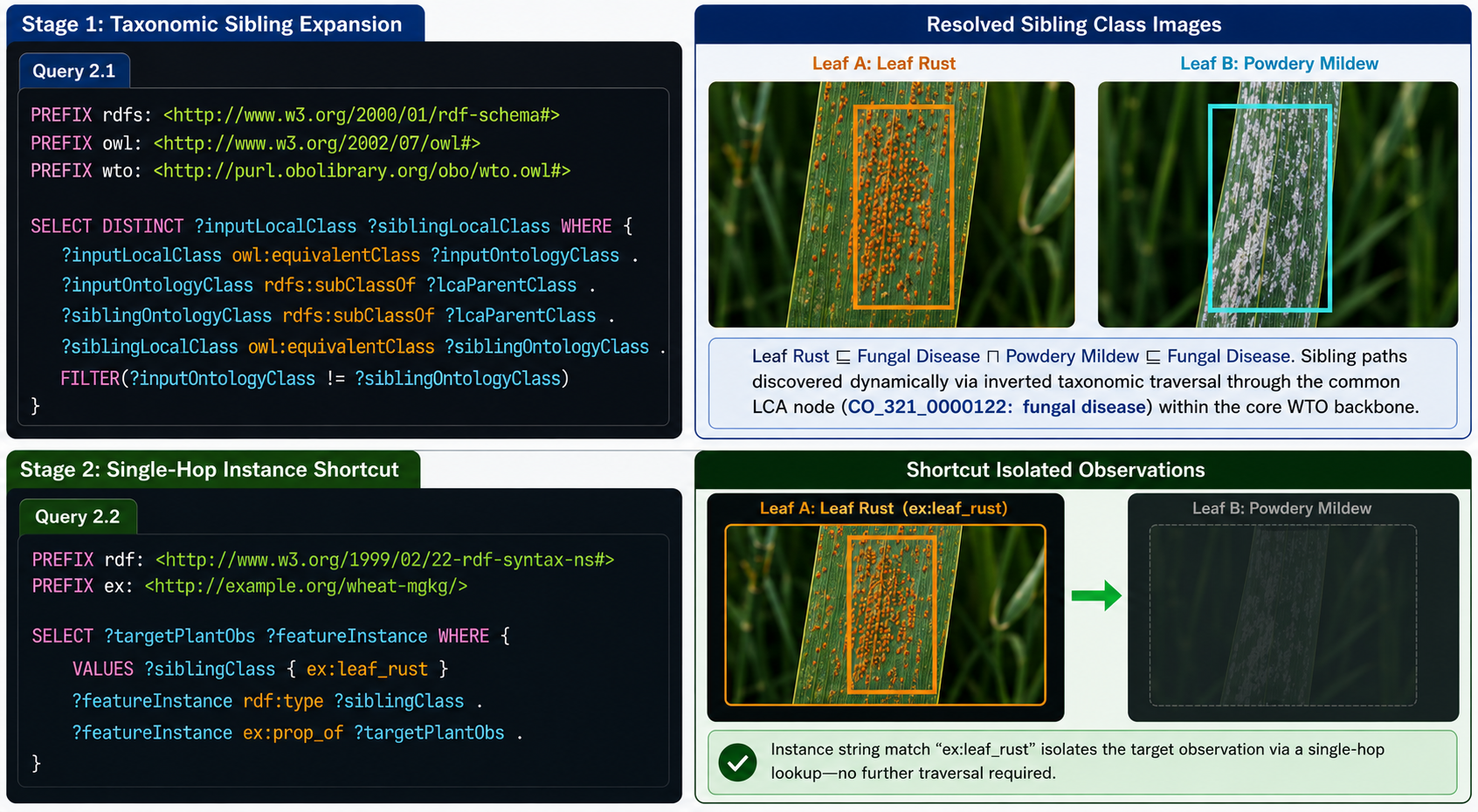}
    \caption{Ontology-subsumption routing tracing inverted taxonomic LCA paths alongside single-hop instance isolation.}
    \label{fig:case_study_2}
\end{figure}

\begin{figure}[H]
\centering
\includegraphics[width=0.8\textwidth]{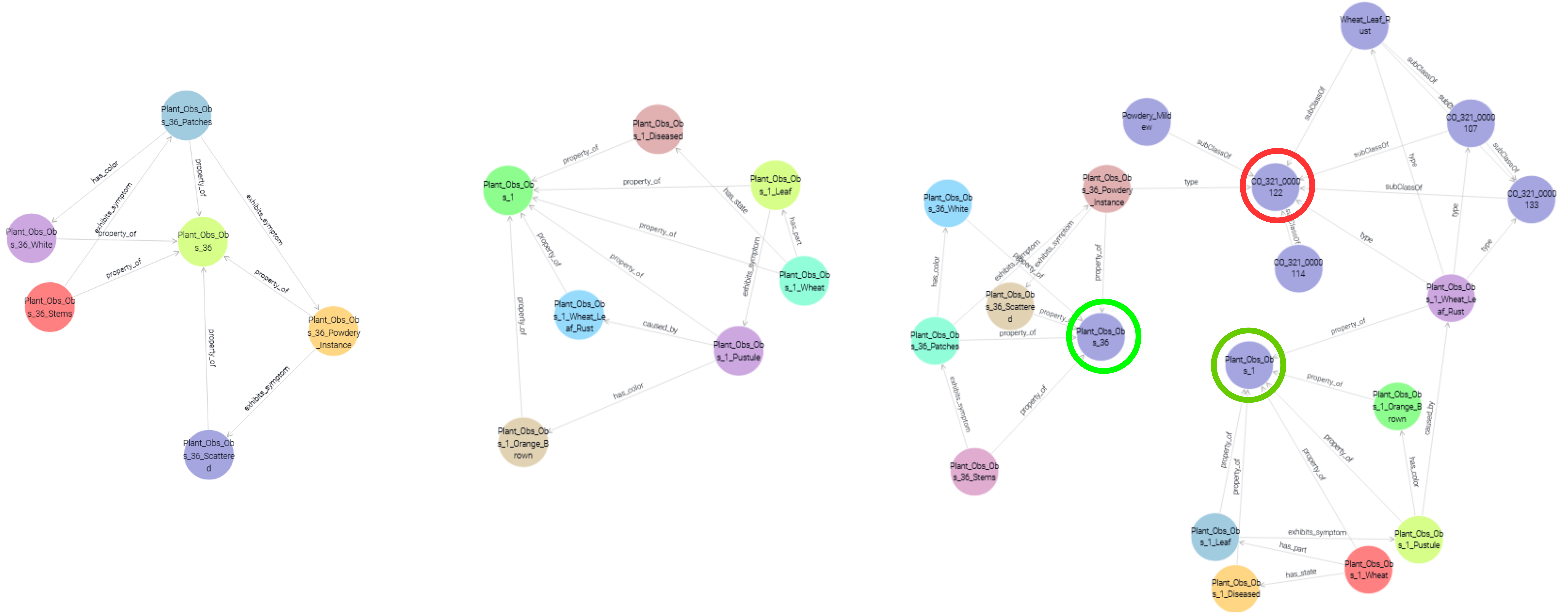}
\caption{Visual representation of the ontology-subsumption and exploratory discovery case study: (Left) Isolated PlantPAD instance neighbourhood tracking powdery mildew; (Center) Isolated WisWheat instance with leaf rust phenotype; (Right) Unified multi-namespace knowledge graph demonstrating converging on the shared canonical node (\texttt{CO\_321\_0000122: fungal disease}) in the WTO backbone.}
\label{fig:knowledge_graph_alignment}
\end{figure}

This performance crash is caused by a hard structural disconnect inherent within the classical hierarchical design of the Wheat Trait Ontology ($\mathcal{WTO}$), where biotic and abiotic paths diverge completely at the highest organizational tiers. Because these domains do not share an actionable, lower-level LCA node within the standard taxonomy, walking up the tree topology to expand a query completely isolates the sibling branches of the opposing domain. The query and  corresponding visual provenance are outlined in Figure~\ref{fig:case_study_2}, while the corresponding neighborhood isolation and final knowledge graph are shown across the sub-panels of Figure~\ref{fig:knowledge_graph_alignment}.

\subsubsection{Automated Downstream Dataset Generation and Quantitative Validation}
\begin{figure}[H]
    \centering
    \includegraphics[width=0.54\linewidth]{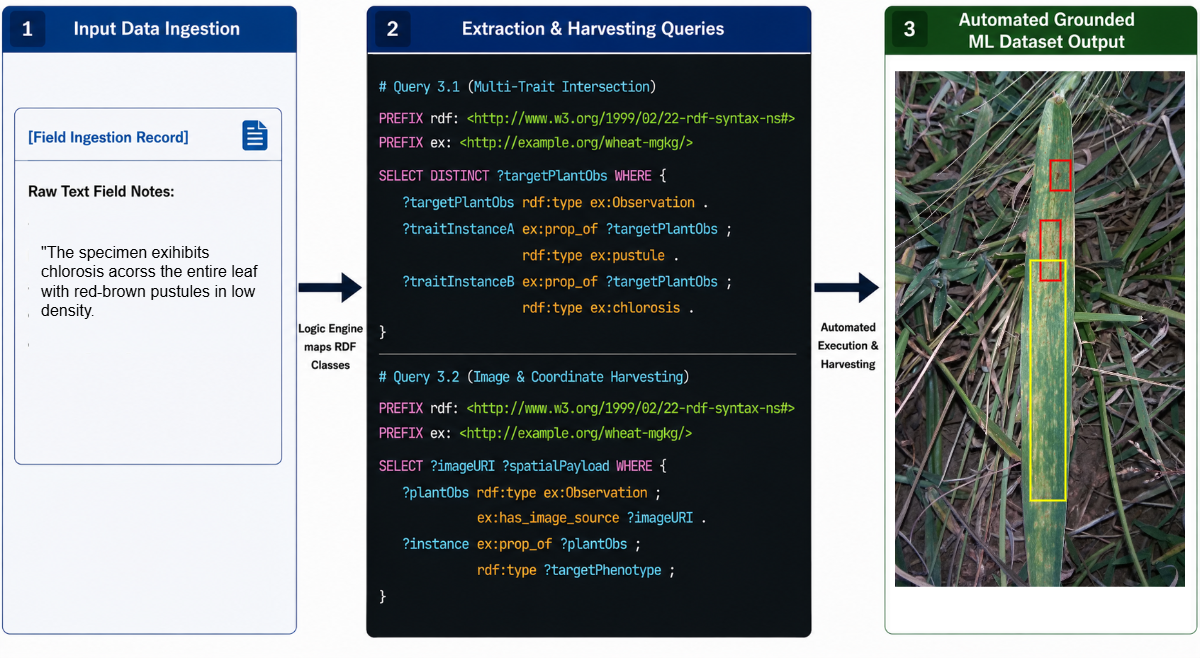}
    \caption{End-to-end automated downstream dataset generation pipeline executing multi-trait logical intersection harvesting.}
    \label{fig:case_study_3}
\end{figure}

\begin{figure}[H]
    \centering
    \includegraphics[width=0.55\textwidth]{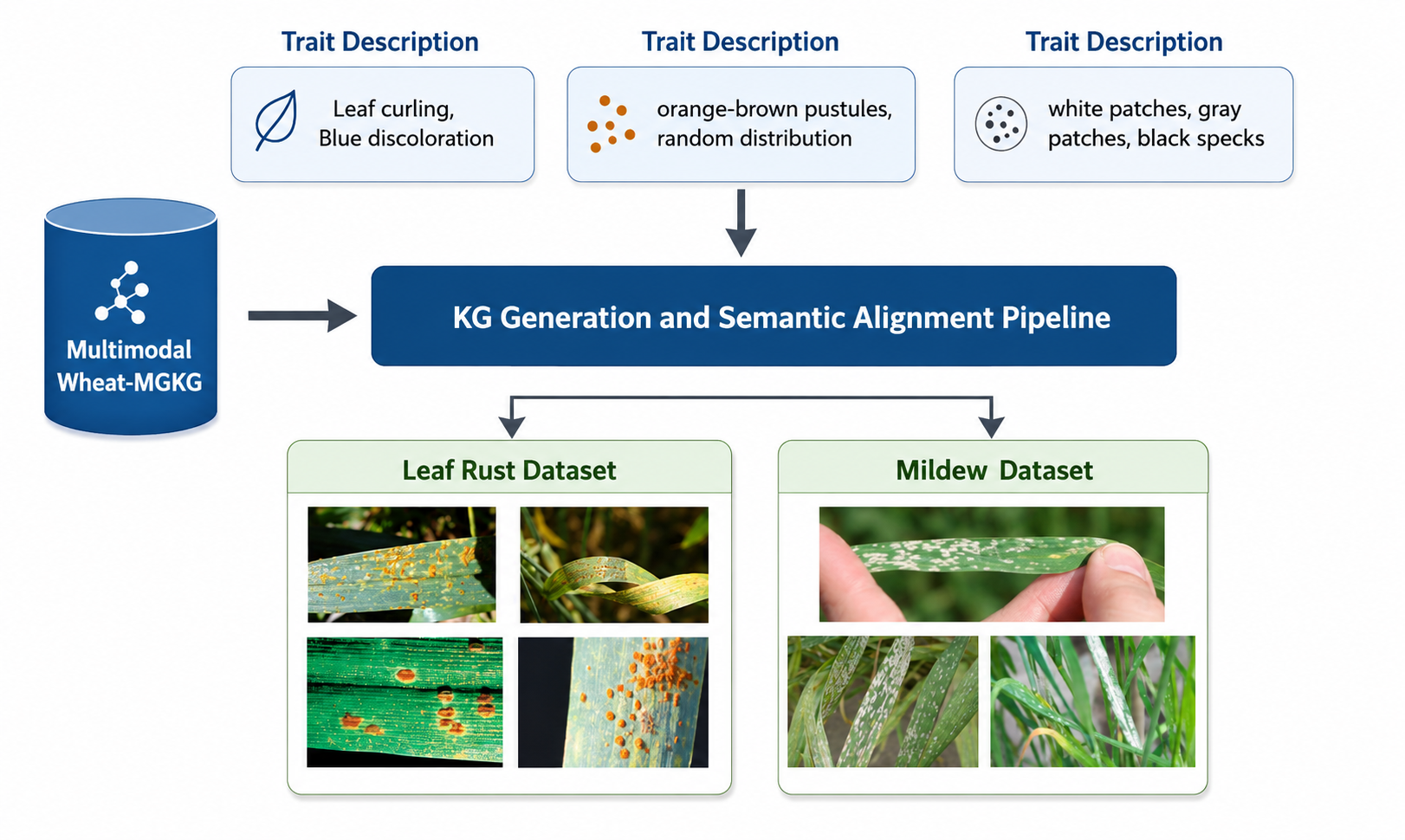}
    \caption{Automated Dataset Curation Validation: System schematic showing text-to-image dataset generation with visual samples of correctly acquired true positives alongside their descriptive query prompts.}
    \label{fig:dataset_builder_proof}
\end{figure}

The quantitative precision of the automated dataset builder was validated across three hidden phenotypic classes, mapping system performance directly against expert human ground truth. Figure~\ref{fig:dataset_builder_proof} provides a schematic of this text-to-image pipeline alongside true positive visual matches, along with the sample query and visual provenance in Figure~\ref{fig:case_study_3}.  Our automated harvesting pipeline and evaluation metrics align with the relation profiling standards established by Lotreck et al. \cite{lotreck2023plant} for constructing plant science knowledge bases. Additionally, the classification errors in our abiotic cohort—where drought symptoms overlap with natural leaf senescence are similar to the phenotyping bottlenecks documented by H{\"u}ther et al. \cite{huther2021aradeepopsis} regarding shared visual signatures across complex leaf states. This phenotypic ambiguity directly validates our use of a strict domain ontology layer to enforce dataset precision without manual sorting overhead. The exhaustive classification results across the selected validation sets are documented in Table~\ref{tab:dataset_curation_metrics}. The framework achieved an \textbf{Overall Framework Accuracy of 78.67\%} while parsing a total of \textbf{2,374 aggregated triples}. 

\begin{table}[H]
\centering
\caption{Quantitative Curation Performance of the Automated Semantic Dataset Builder Across Target Phenotypic Classes}
\label{tab:dataset_curation_metrics}
\setlength{\tabcolsep}{5pt}
\renewcommand{\arraystretch}{1.05}
\begin{tabular}{lcccccc}
\toprule
\textbf{Target Phenotypic Cohort} & \textbf{True Pos.\ (TP)} & \textbf{False Neg.\ (FN)} & \textbf{False Pos.\ (FP)} & \textbf{Precision} & \textbf{Recall} & \textbf{F1-Score} \\
\midrule
Leaf Rust       & 18 & 7 & 2 & 90.00\%  & 72.00\% & 0.80 \\
Powdery Mildew  & 19 & 6 & 2 & 90.48\%  & 76.00\% & 0.83 \\
Drought Stress  & 15 & 8 & 3 & 83.33\%  & 65.22\% & 0.73 \\ 
Healthy Control & 22 & 3 & 0 & 100.00\% & 88.00\% & 0.94 \\
\midrule
\multicolumn{7}{l}{\textbf{Overall Dataset Curation Metrics:}} \\
\multicolumn{4}{l}{~~$\bullet$ Overall Framework Accuracy: \textbf{74.47\%}} & \multicolumn{3}{l}{~~$\bullet$ Total Aggregated Triples: 2,374} \\
\multicolumn{4}{l}{~~$\bullet$ Total Evaluated Samples: 98} \\
\bottomrule
\end{tabular}
\end{table}
The results in Table~\ref{tab:dataset_curation_metrics} show that the automated pipeline works reliably on its own. The engine performed best on the \textit{Healthy Control} group, reaching perfect precision ($100.00\%$) and high recall ($88.00\%$). This is because healthy wheat leaves have clear, uniform shapes and colors, making it easy for the Vision-Language Model (VLM) to identify them without getting confused by background noise.

In contrast, the system missed some instances of the biotic diseases, yielding a recall of $72.00\%$ for \textit{Leaf Rust} and $76.00\%$ for \textit{Powdery Mildew}. The lowest performance layer occurred when confronting the abiotic cohort, where \textit{Drought Stress} dropped to a recall of $65.22\%$ alongside a lower precision of $83.33\%$. In a real field setting, abiotic stress expressions, such as general leaf rolling, tipping, or localized drying frequently overlap with natural plant senescence or background structural anomalies. This phenotypic ambiguity might cause the early-stage vision backend to drop initial trait extractions or generate false-positive boundary leaks.

However, the framework still achieved strong precision values across the disease groups ($90.00\%$ for Leaf Rust and $90.48\%$ for Powdery Mildew). This directly proves the value of our neuro-symbolic setup: even if the vision engine struggles with a challenging or early-stage symptom, the system successfully blocks false matches by double-checking the prospective relations against the structured rules within the Wheat Trait Ontology ($\mathcal{WTO}$) backbone. By traversing a single-hop shortcut (\texttt{ex:property\_of}) back to the centralized $Plant\_Obs\_Id$ anchor node, the pipeline harvests highly accurate training datasets automatically and completely removes the need for manual data sorting.
\subsection{Operational Workflow and UI Usage Demonstration}

\begin{figure}[H]
    \centering
    \begin{subfigure}[b]{0.49\textwidth}
        \centering
        \includegraphics[width=\textwidth]{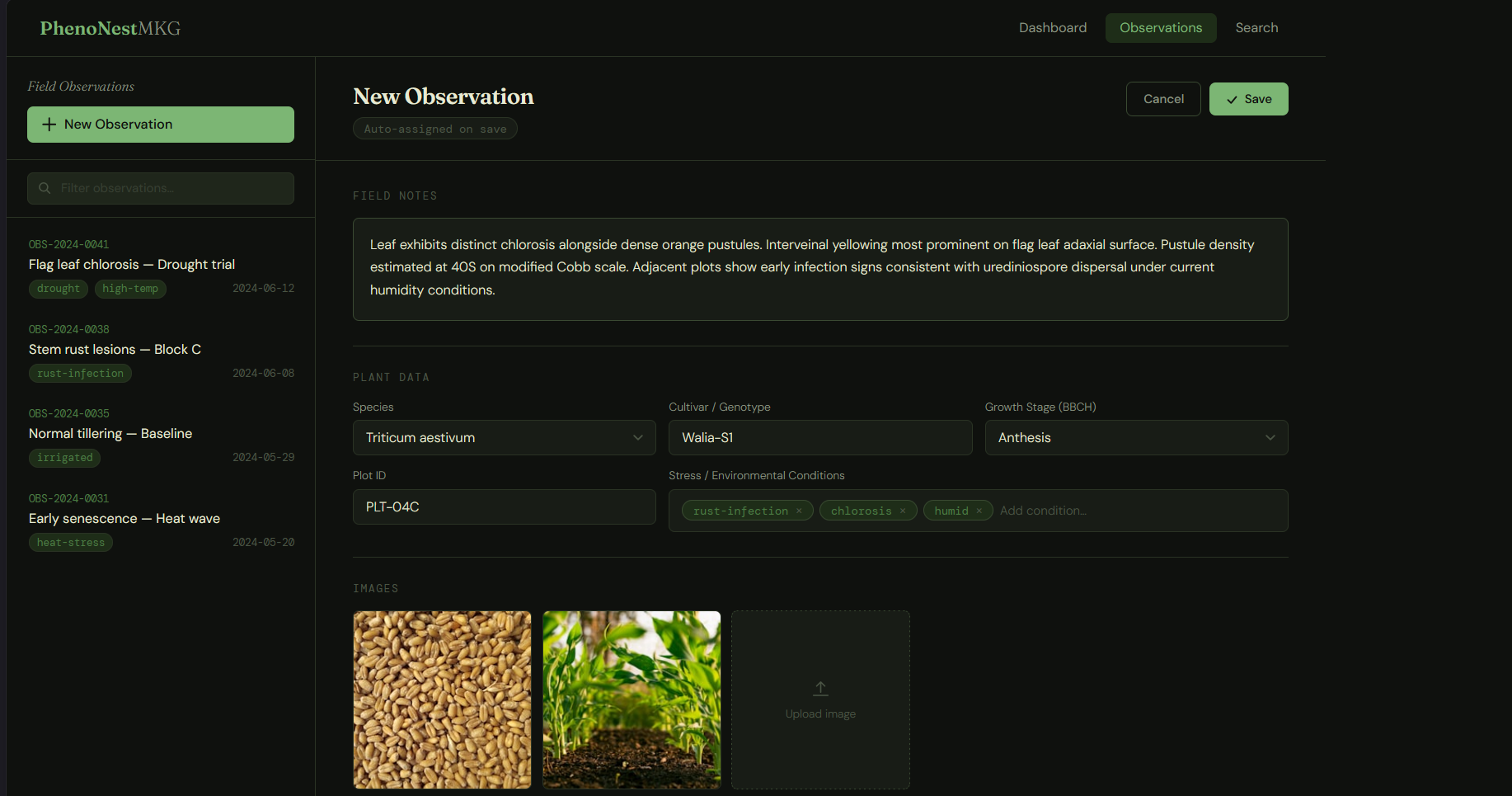}
        \caption{Screen 1 initialization state for loading new field specimen payloads.}
        \label{fig:wf_new_obs}
    \end{subfigure}
    \hfill
    \begin{subfigure}[b]{0.49\textwidth}
        \centering
        \includegraphics[width=\textwidth]{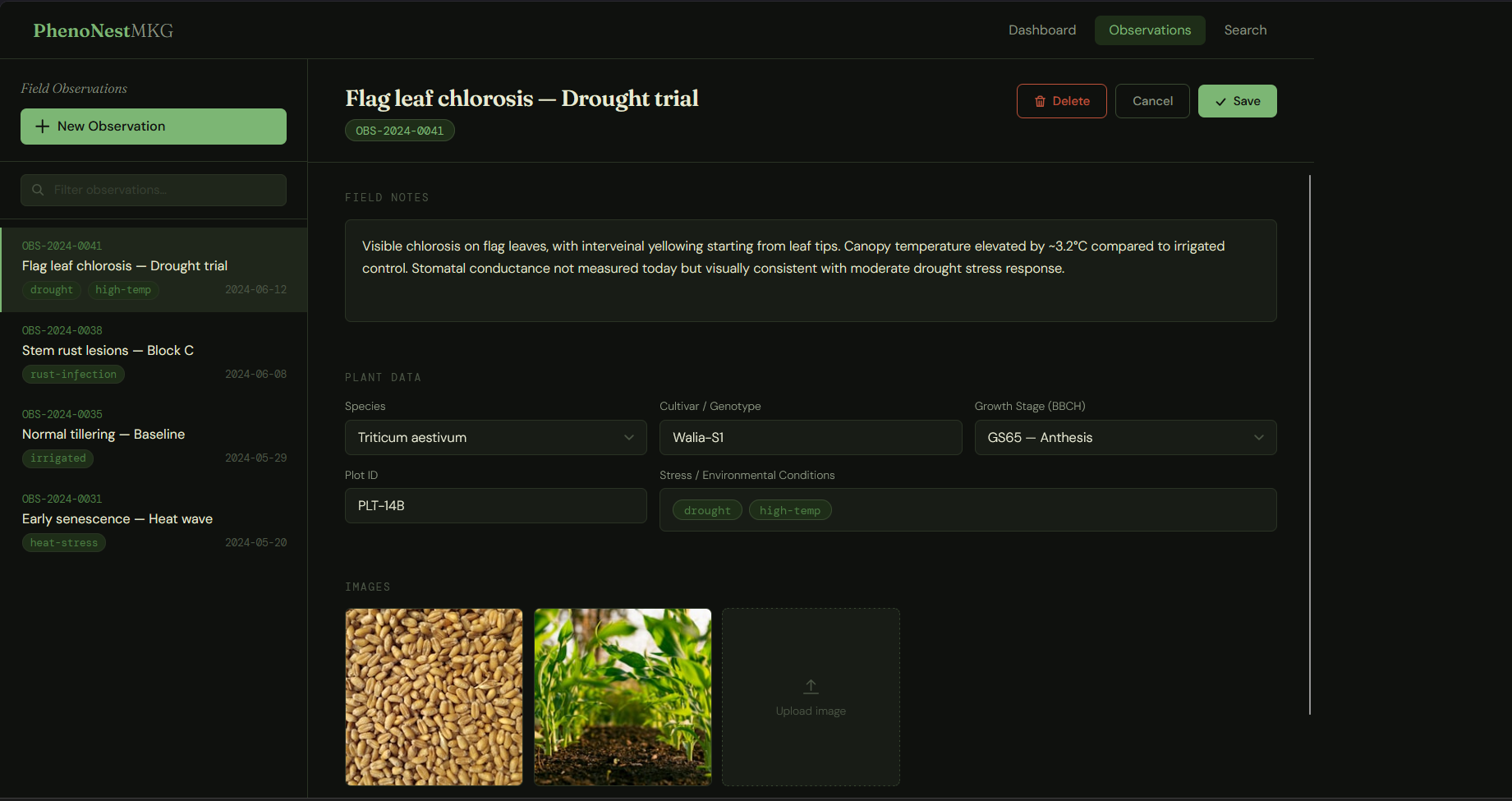}
        \caption{Screen 1 master-detail loaded state for an individual observation$\mathcal{B}$.}
        \label{fig:wf_sel_obs}
    \end{subfigure}
    
    \vspace{0.3cm} 
    
    \begin{subfigure}[b]{0.49\textwidth}
        \centering
        \includegraphics[width=\textwidth]{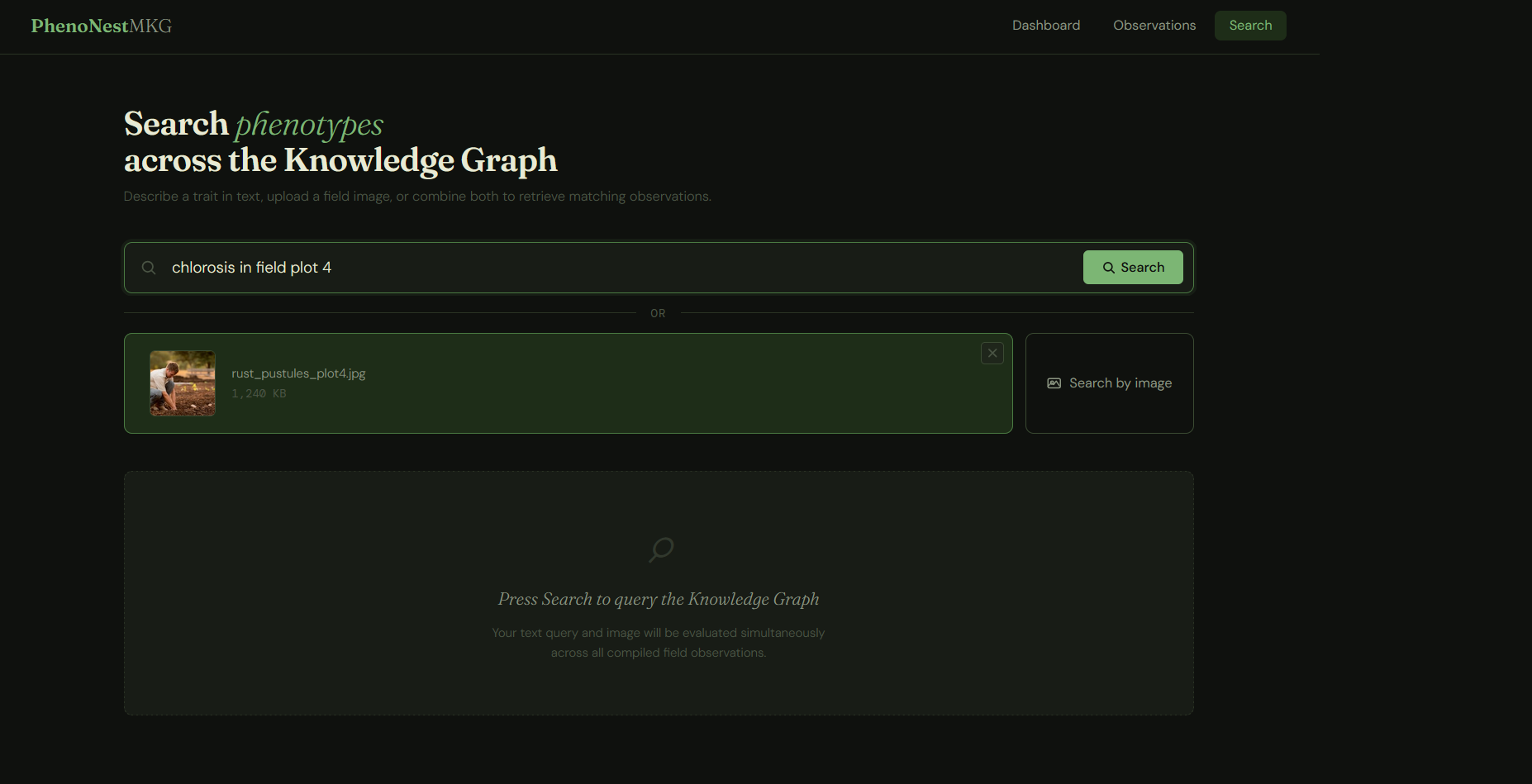}
        \caption{Screen 2 interface for concurrent text and drag-and-drop image hybrid queries.}
        \label{fig:wf_search_in}
    \end{subfigure}
    \hfill
    \begin{subfigure}[b]{0.49\textwidth}
        \centering
        \includegraphics[width=\textwidth]{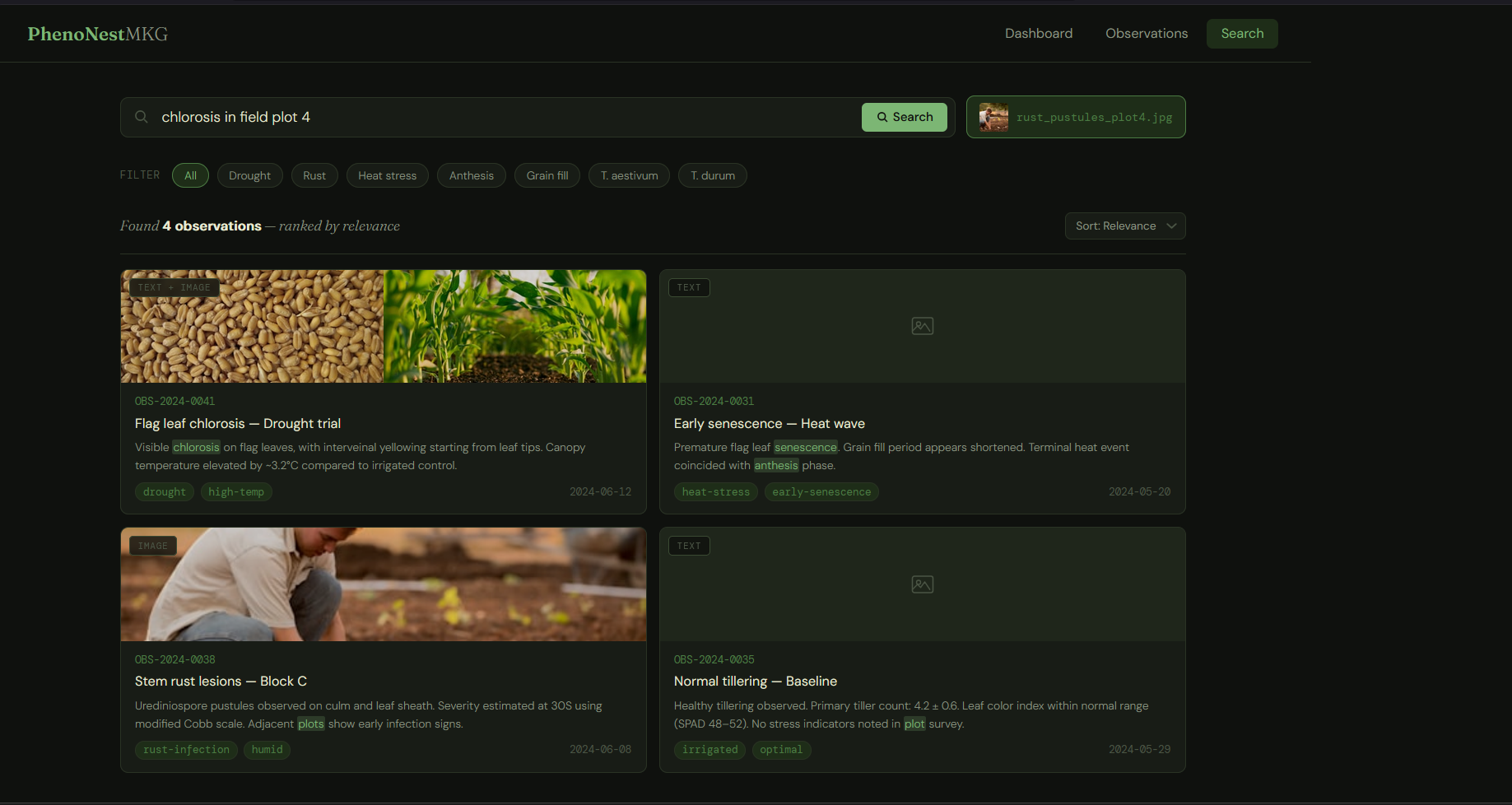}
        \caption{Screen 2 output view rendering metadata badges.}
        \label{fig:wf_search_res}
    \end{subfigure}
    
    \caption{Step-by-step user workflow and interactive dashboard states showing data curation card inputs (a), master-detail entity retrieval (b), hybrid query parameters (c), and cross-modal semantic search results (d).}
    \label{fig:ui_workflow_lifecycle}
\end{figure}

To illustrate the practical utility of the interface in active field-research scenarios, we outline a typical user interaction lifecycle mapping a new field observation from ingestion to multimodal discovery:

\begin{enumerate}
    \item \textbf{Data Ingestion and Annotation (Screen 1):} When a researcher identifies an anomalous plant state in the field, they open the \textit{Observation Manager}. Clicking the ``New Observation'' button initializes a fresh administrative record. The user uploads a high-resolution leaf canopy image and inputs raw field notes into the text area (e.g., \textit{``Leaf exhibits distinct chlorosis alongside dense orange pustules''}). Using the tag-style interactive fields shown in Figure~\ref{fig:framework_ui_dashboards}a, the researcher quickly updates the structural fields for cultivar, plot ID, and BBCH growth scale. Clicking ``Save'' triggers an asynchronous \texttt{POST} request, executing the backend text distillation and RDF-typing logic to formally update the centralized \texttt{Plant\_Obs\_Id} node in the active knowledge graph database.
    \item \textbf{Master-Detail Navigation (Screen 1):} For historical verification or record modification, the researcher can utilize the filterable sidebar layout in Figure~\ref{fig:framework_ui_dashboards}a. Selecting an arbitrary \texttt{Plant\_Obs\_Id} immediately updates the right-hand dashboard window via local state management. This loads the existing field note segments, populates the bound ontology tags mapped by PlantDeBERTa, and displays the image grid along with its spatial grounding bounding boxes ($\mathcal{B}$) to visually verify the recorded traits.
    \item \textbf{Multimodal Knowledge Discovery (Screen 2):} To perform downstream exploratory search across compiled experiments, the researcher navigates to the \textit{Multimodal Search Interface} (Figure~\ref{fig:framework_ui_dashboards}b). If searching by visual criteria, the user drags an image showing questionable rust pustules into the designated drop zone. Concurrently, they can type a text query like \textit{``chlorosis in field plot 4''} to execute a hybrid search. As shown in the layout, the system evaluates the text embeddings and visual features simultaneously, outputting a prioritized stack of result cards decorated with relevance match scores, highlighted text snippets. Clicking any returned search result instantly navigates the user back to the corresponding master-detail card in Screen 1 for deeper analysis.
 The overall operational workflow has been shown in Figure~\ref{fig:ui_workflow_lifecycle} from data curation dashboard to hybrid search.
\end{enumerate}
\section{Conclusions}

In this work, we presented PhenoNest-MKG, a novel neuro-symbolic framework designed to bridge the long-standing semantic gap between high-resolution field imagery and unstructured biological field notes. By explicitly separating visual perception from semantic reasoning into a decoupled two-engine pipeline, the system ensures architectural modularity and precision in high-throughput phenotyping workflows. The introduction of a centralized observation node (\texttt{Plant\_Obs\_Id}) establishes a robust temporal anchor, seamlessly linking disparate multimodal subgraphs across longitudinal tracking intervals and distinct field experiments.

This architecture effectively resolves key research limitations by dismantling fragmented agricultural data silos and eliminating structural incoherence across specialized datasets. Moving away from traditional vision-dominant frameworks that provide only a coarse mapping between visual features and text descriptions, PhenoNest-MKG establishes a mathematically traceable neuro-symbolic pathway. This pipeline successfully anchors abstract symbolic knowledge and domain ontologies directly to the absolute physical pixels of the crop canopy layout.

Despite these architectural contributions, several operational limitations remain to be addressed in future research. The system's current dependency on absolute bounding-box token sequences during generative visual grounding introduces decoding latency, which increases computational overhead and may constrain real-time deployment on field-edge processing hardware. Future horizons will focus on optimizing this coordinate projection layer to enable real-time execution a major direction to explore will be the training of models smaller VLMs that optimize the vision backbone to reduce parameter count drastically. Furthermore, we aim to scale the logic engine to model multi-stressor predictive disease kinetics over extended crop life cycles and deploy cross-ontology mapping utilities to extend the framework's generalization capabilities to other staple cereal crops.

\section*{Acknowledgements}
The authors gratefully acknowledge the computational infrastructure and resources provided by the Centre of Studies in Resources Engineering (CSRE) at the Indian Institute of Technology Bombay.

\section*{CRediT authorship contribution statement}
\textbf{Jayant Ghadge:} Methodology, Software, Data curation, Visualization, Investigation, Validation, Formal analysis, Writing – original draft, Writing – review \& editing. 
\textbf{Soumyashree Kar:} Conceptualization, Methodology, Visualization, Investigation, Supervision, Validation, Formal analysis, Writing – original draft, Writing – review \& editing, Project administration.
\textbf{Surya S. Durbha} Supervision, Validation, Formal analysis, Writing – review \& editing, Project administration.

\section*{Declaration of competing interest}
The authors declare that they have no known competing financial interests or personal relationships that could have appeared to influence the work reported in this paper.

\section*{Data availability}
The data used in this study are derived from publicly available, open-source repositories (including WisWheat, GWHD, GWFSS, WFD2020, and TAEC). All relevant data sources, repository links, and dataset characteristics are explicitly detailed and cited within the 'Data Curation' section of the manuscript.

\section*{Code availability}
The code developed for the PhenoNest framework will be made available upon reasonable request to the corresponding author.

\bibliographystyle{unsrtnat}
\bibliography{refrences.bib}

\end{document}